\documentclass[journal]{new-aiaa}
\usepackage[utf8]{inputenc}
\usepackage{textcomp}

\usepackage{amsmath,amssymb,mathrsfs,framed,esint,slashed,url}

\usepackage{siunitx}

\usepackage{array} 
\usepackage{placeins} 
\usepackage{subcaption} 
\usepackage{nicematrix} 
\usepackage{graphicx}
\usepackage{color}
\usepackage{xcolor} 
\usepackage[normalem]{ulem} 
\usepackage[all]{xy}
\usepackage{booktabs}
\usepackage{hyperref}
\hypersetup{
    bookmarksnumbered=true,
    bookmarkstype=toc
}

\newcommand{\rem}[1]{}

\newcommand{\bOmega}{{\boldsymbol{\Omega}}}

\newcommand{\beq}{\begin{equation}}
\newcommand{\eeq}{\end{equation}}
\newcommand{\bal}{\begin{align}}
\newcommand{\eal}{\end{align}}

\title{Estimation of Spacecraft Inertia Tensor Using Attitude-Only Data from Torque-Free Motion} 
\author{
    Daigo Kobayashi\footnote{
    Assistant Professor, Department of Aerospace Engineering and Mechanics, 
    213 Hardaway Hall, Tuscaloosa, Alabama 35487-0350, USA. 
    }
    and 
    Vakhtang Putkaradze\footnote{
    Professor, Department of Mathematics, 
    345 Gordon Palmer Hall, Tuscaloosa, Alabama 35487-0350, USA. 
    Corresponding author: vputkaradze@ua.edu
    }
}
\affil{The Unviersity of Alabama, Tuscaloosa, Alabama 35487, USA}

\begin{document}

\maketitle

\thispagestyle{plain}
\pagestyle{plain}

\begin{abstract}
We present an attitude-only framework for estimating a spacecraft’s normalized inertia tensor from torque-free rotational motion. Our method supports both continuous single-arc observations and the joint use of multiple short torque-free arcs, while requiring neither gyroscope measurements nor known control torques. A Karush–Kuhn–Tucker formulation provides a fast linear initialization, which is refined by nonlinear shooting using the exact Jacobi-elliptic solution of Euler’s equations and a Magnus-expansion quaternion map. Under controlled attitude noise, tests using a single 500-second arc reduced inertia-tensor error by approximately one order of magnitude relative to an Extended Kalman Filter initialized from the same estimate, while requiring nearly two orders of magnitude less computation. Joint estimation from three 100-second arcs provided a similar improvement in accuracy and remained more than one order of magnitude faster. Photorealistic proximity-operations simulations further evaluated both strategies using monocular image-derived attitudes. The 2,000-second single-arc cases achieved sub-thousandth median inertia-tensor error and supported 10-hour attitude predictions with single-digit-degree median error. In three-arc cases using 30–300 seconds per arc, our method consistently outperformed the EKF refinement, with performance governed by rotational excitation and temporal sampling. 
\end{abstract}

\section{Introduction}
\subsection{Background and Previous Work}
\lettrine{T}{he} inertia tensor is a fundamental parameter in spacecraft rotational dynamics. It determines how a spacecraft responds to internal and external torques and is therefore essential for attitude prediction, maneuver design, and control-system implementation. Accurate inertia knowledge is also important for estimating propellant consumption, monitoring structural changes caused by fuel depletion or appendage deployment, and assessing the rotational state of inoperative spacecraft during rendezvous, docking, recovery, or debris-removal operations \cite{geitgey2006determination}. For non-cooperative targets, in particular, accurate inertia estimates enable long-term prediction of tumbling motion, preferably with minimal onboard computation and without requiring active excitation.

Existing inertia-identification methods typically rely on gyroscope measurements, known control torques, or both. Batch least-squares estimation was demonstrated for the Cassini spacecraft using flight data \cite{lee2002flight}, while extended linear regression has been used to account for noisy angular velocity and torque measurements \cite{jun2010identification}. Extended Kalman Filter (EKF) formulations have also been widely used for telemetry-based parameter estimation, including approaches based on angular momentum and rotational kinetic energy rather than direct numerical differentiation of angular velocity \cite{norman2011orbit}. Related work has considered robotic platforms and UAVs \cite{nabavi2016unmanned,muliadi2017estimating}, variational-integrator formulations with physical inertia constraints \cite{keim2006spacecraft,manchester2017recursive}, Unscented Kalman Filters  (UKFs) \cite{linares2012space}, two-stage EKF/RLS estimation \cite{yang2015new}, and Savitzky-Golay filtering followed by EKF refinement \cite{kim2016rigid}.  More recently, Candan and Servadio \cite{Candan_2026} developed an augmented UKF that jointly estimates relative kinematic states and the normalized inertia tensor of an uncooperative spacecraft by fusing monocular keypoint detections with LiDAR-assisted depth measurements; they reported convergence within 500 s and steady-state standard deviations below 0.05\% for the diagonal inertia components for long-term observations. 

Several studies have addressed special cases, observability, or robustness. For axisymmetric bodies, simplified torque-free dynamics admit analytical mass-property estimation \cite{ma2017estimation}. Other work has used observation-based optimization \cite{ivanov2018relative}, EKF estimation with gyroscope data and known torques \cite{bellar2019satellite}, and combined star-tracker, gyroscope, and control-mechanism information \cite{cheriet2021inertia}. Since EKF-based estimates can be sensitive to data quality and biased in the off-diagonal inertia terms, instrumental-variable methods have been proposed as alternatives \cite{nainer2018orbit,nainer2019flight}, including active experiment design to improve inertia observability \cite{nainer2020design}. Additional developments include comparisons of Instrumental Variable (IV) and UKF observability \cite{magnani2022satellite}, finite-time concurrent-learning identification from stored telemetry \cite{zhao2020finite}, robust coupled least-squares estimation for time-varying inertia during on-orbit assembly \cite{ni2024identification}, and zonotopic filtering for bounded-noise parameter estimation \cite{wang2025orbit}. 

A particularly relevant formulation was introduced in \cite{tanygin1997mass}, where the rotational equations were projected onto the angular velocity vector to obtain a linear relation between actuator work and the inertia tensor. In the torque-free case, this relation becomes homogeneous and can, in principle, still constrain the normalized inertia tensor. However, such approaches remain strongly dependent on accurate angular velocity information. This dependence is problematic for small spacecraft equipped with low-cost MEMS gyroscopes and is prohibitive for non-cooperative space-domain-awareness scenarios, where direct rate telemetry is unavailable. In contrast, attitude histories may be available from onboard star trackers or extracted externally from ground- or space-based imagery \cite{zarei2018motion}. Mason et al. \cite{mason2023learning} considered a more direct use of image sequences by introducing a physics-informed neural network that learns an $\mathrm{SO}(3)$-structured latent representation and an inertia tensor from image sequences to predict rigid-body rotational dynamics. Although evaluated primarily for future-image prediction rather than inertia-estimation accuracy, their work provides a closely related example of learning rotational dynamics directly from images.

\subsection{Focus of This Paper: Estimating Inertia Tensor from Attitude Data for Free Motion}
This paper addresses the problem of estimating a spacecraft's normalized inertia tensor from attitude histories alone during torque-free rotational motion. The proposed framework eliminates the need for gyroscope measurements, known control torques, or active excitation maneuvers. Instead of reconstructing angular velocity by differentiating noisy quaternion data, the method analytically propagates the underlying rotational dynamics using the exact torque-free rigid-body solution and a quaternion mapping based on the Magnus expansion, thereby reducing sensitivity to measurement noise. 

The proposed estimator consists of a fast Karush-Kuhn-Tucker initialization followed by nonlinear shooting refinement. The initialization provides a linear estimate of the normalized inertia tensor, while the refinement minimizes attitude mismatch using the analytical phase flow of Euler's equations. This structure avoids the iterative matrix-propagation overhead of conventional EKF-based approaches and is therefore suitable for efficient onboard computation.

The framework is also evaluated under realistic image-derived attitude errors relevant to rendezvous and proximity operations. Attitude measurements are recovered from monocular images by detecting the image locations of predefined body-frame keypoints with deep neural networks and solving the resulting 2D-3D pose-estimation problem. The resulting attitude errors are generally non-Gaussian and depend on spacecraft geometry, illumination, and relative viewing direction. By demonstrating inertia estimation from passive optical attitude histories, this work connects torque-free mass-property identification with image-based pose estimation for non-cooperative spacecraft. The results show that accurate inertia estimation is possible when the rotational motion provides sufficient directional excitation, while also identifying degraded observability near principal-axis rotations.

\section{Analytical Phase-Space Mapping and Attitude Kinematics}
\label{sec:analytical_mapping}

This section outlines the two-tier mathematical framework used to propagate the spacecraft's state over a discrete time interval $h = t_{k+1} - t_k$. This derivation follows the classical Jacobi-elliptic-function solution of torque-free rigid-body motion \cite[Chapter~5.7]{lawden2013elliptic}; see also \cite{whittaker1964treatise}. First, we leverage the exact, closed-form solutions of Euler's equations via Jacobi elliptic functions to map the body-frame angular velocities. Second, because a closed-form solution for the corresponding attitude kinematics does not exist for asymmetric rigid bodies, we apply the Magnus expansion to resolve the transition matrix for the attitude quaternion.

\subsection{Body-Frame Angular Velocity Propagation via Jacobi Elliptic Functions}

To achieve a highly precise trajectory map without the truncation errors inherent in discrete numerical solvers (such as explicit Runge-Kutta schemes), the angular velocity vector is propagated analytically across the step interval. 


In a torque-free environment, the rigid-body dynamics are governed by the extended Euler equations for a rigid top, tracking the body-frame angular velocity $\boldsymbol{\Omega}$ and attitude $g\in SO(3)$:
\begin{equation}
\mathbf{I}\dot{\boldsymbol{\Omega}}(t)
+
\widehat{\Omega}(t)\mathbf{I}\boldsymbol{\Omega}(t)
=
\mathbf{0},
\quad
\dot{g}
=
\widehat{\Omega}g.
\label{ext_Euler_eq}
\end{equation}
Here, $\mathbf{I}\in\mathbb{R}^{3\times3}$ is the symmetric positive-definite inertia tensor, and $\widehat{\Omega}:=\boldsymbol{\Omega}^{\times}\in\mathfrak{so}(3)$ denotes the skew-symmetric matrix satisfying $\widehat{\Omega}\mathbf{x}=\boldsymbol{\Omega}\times\mathbf{x}$ for all $\mathbf{x}\in\mathbb{R}^3$. Instead of working with the $3\times3$ orthogonal attitude matrix $g$, we use the unit quaternion representation $\mathbf{q}=(q_0,q_1,q_2,q_3)$, with $\|\mathbf{q}\|=1$~\cite{altmann2005rotations}.

This system is constrained by two fundamental algebraic invariants of motion representing kinetic energy $T$ and the square of the angular momentum magnitude $L^2$:

\begin{equation}
    E = \frac{1}{2} \boldsymbol{\Omega} \cdot  \mathbf{I} \boldsymbol{\Omega} = \text{constant (Energy)}, \quad 
    L^2 = \left\| \mathbf{I}\boldsymbol{\Omega} \right\|^2  = \text{constant (Casimir)}
\end{equation}

The intersection of these energy and momentum ellipsoids defines the geometric polhode path tracking the evolution of $\boldsymbol{\Omega}(t)$. The analytical solution partitions into distinct topological regimes relative to the intermediate principal axis of inertia ($I_2$):
\begin{itemize}
    \item \textbf{Case 1 (Major Axis Orbit):} $L^2 > 2T I_2$, where the state circulates around the axis of maximum inertia.
    \item \textbf{Case 2 (Minor Axis Orbit):} $L^2 < 2T I_2$, where the state circulates around the axis of minimum inertia.
\end{itemize}


Given the initial angular velocity $\boldsymbol{\omega}(0)=\boldsymbol{\Omega}_0$, the torque-free Euler equations admit an exact analytical propagation over a time step $h$. We denote this exact angular-velocity flow map by
\begin{equation}
\boldsymbol{\Omega}(h)
=
\mathcal{F}(\boldsymbol{\Omega}_0,\mathbf{I},h),
\label{omega_exact_map}
\end{equation}
where $\mathcal{F}(\cdot)$ is evaluated using the classical Jacobi-elliptic-function solution of Euler's equations. This map returns the angular velocity at time $h$ without numerical integration of the rotational dynamics. The explicit construction of $\mathcal{F}$, including the elliptic modulus, characteristic frequency, and amplitude coefficients, is given in Appendix~\ref{app:euler_derivation}.

\subsection{Scale Invariance and Normalization}

The torque-free rotational dynamics are invariant under a uniform scaling of the inertia tensor,
\begin{equation}
\mathbb{I} \rightarrow \alpha \mathbb{I}, \qquad \alpha > 0 .
\end{equation}
Consequently, attitude-only observations cannot determine the absolute scale of the inertia tensor; they can only identify its relative mass distribution. To remove this scalar ambiguity, we estimate the normalized inertia tensor by imposing the unit-trace constraint
\begin{equation}
\operatorname{tr}\mathbb{I}=1 .
\label{I_norm_choice}
\end{equation}
This normalization treats all coordinate axes symmetrically and avoids the arbitrary choice of fixing a particular tensor component, such as $I_{11}=1$.

\subsection{Approximation of the Attitude through the Magnus Expansion}

While the angular velocity $\boldsymbol{\Omega}(t)$ is exactly integrable, the kinematic differential equation governing the attitude quaternion $\mathbf{q}(t) = [q_0, q_1, q_2, q_3]^T$ is generally non-integrable in closed form for fully asymmetric bodies. The evolution of quaternion, replacing the attitide equation in \eqref{ext_Euler_eq} is given by the linear time-dependent system:
\begin{equation}
    \dot{\mathbf{q}}(t) = \frac{1}{2} \overline{\Omega}_K(t) \mathbf{q}(t)\, , 
\end{equation}
where $\overline{\Omega}_K(t)$ is the $4 \times 4$ skew-symmetric matrix:
\begin{equation}
    \overline{\Omega}_K = \begin{bmatrix} 
    0 & -\Omega_x & -\Omega_y & -\Omega_z \\ 
    \Omega_x & 0 & \Omega_z & -\Omega_y \\ 
    \Omega_y & -\Omega_z & 0 & \Omega_x \\ 
    \Omega_z & \Omega_y & -\Omega_x & 0 
    \end{bmatrix}
\end{equation}
Let us, for the simplicity of the notation, assume that the starting point is at $t=0$ and end point is at $t=h$. The case when the initial point is at $t=t_k$ and end point is at $t=t_k+h = t_{k+1}$ is considered similarly. 

Because the kinematic matrix $\mathbf{\Omega}_K(t)$ at different time points do not generally commute, (i.e., $[\mathbf{\Omega}_K(t_1), \mathbf{\Omega}_K(t_2)] \neq \mathbf{0}$), one cannot use a simple matrix exponential solution. To resolve this non-commutativity, we represent the solution using the Magnus expansion applied to quaternions:
\begin{equation}
    \mathbf{q}(h) = \exp\left( \mathbf{\Theta}(h) \right) \mathbf{q}(0)
\end{equation}
where the matrix exponent $\mathbf{\Theta}(h)$ is defined as an infinite series of nested commutators:
\begin{equation}
    \mathbf{\Theta}(h) = \sum_{m=1}^{\infty} \mathbf{\Theta}_m(h)
\end{equation}
The first few terms of this series, which capture the non-commutative corrections, are given by:
\begin{equation}
\begin{aligned}
\widehat{\Theta}_1(h) &= \frac{1}{2} \int_{0}^{h} \widehat{\Omega}_K(t_1)  dt_1 \, , \\
\widehat{\Theta}_2(h) &= \frac{1}{8} \int_{0}^{h} 
\int_{0}^{t_1}   [\widehat{\Omega}_K(t_1), \widehat{\Omega}_K(t_2)]  dt_1 dt_2 \, , \, \ldots 
\end{aligned}
\label{Magnus_expansion}
\end{equation}
By truncating this series at a specific order $m$ (in our case, $m=2$), one 
constructs a numerical integrator that is both symplectic (in the context of 
the Lie group) and unconditionally stable. This allows for larger time steps 
$h$ while maintaining the geometric properties---specifically the 
unit-norm constraint---of the quaternions representing the rigid body's motion.

The first-order term $\widehat{\Theta}_1(h)$ in \eqref{Magnus_expansion} 
represents the mean angular displacement over the step interval, while the 
second-order term $\widehat{\Theta}_2(h)$ introduces the leading 
non-commutative correction to account for the shifting axis of rotation. 

We approximate the angular velocity matrix $\widehat{\Omega}_K(t)$ over the 
interval $t \in [0,h]$ as a linear function of time. Under this linear 
assumption, the integrals in \eqref{Magnus_expansion} evaluate analytically. 
This provides an exact quadrature mapping that connects the initial quaternion 
$\mathbf{q}(0)$ to the final quaternion $\mathbf{q}(h)$ based on the profile 
of $\boldsymbol{\Omega}(t)$. Consequently, when the discrete angular velocity 
trajectory $(\boldsymbol{\Omega}_0, \boldsymbol{\Omega}_1, \ldots, 
\boldsymbol{\Omega}_N)$ is determined exactly via the analytical map 
\eqref{omega_exact_map}, the corresponding quaternion sequence $(\mathbf{q}_0, 
\mathbf{q}_1, \ldots, \mathbf{q}_N)$ can be resolved to  high 
precision using the Magnus expansion. 

Under the assumption that the angular velocity matrix $\widehat{\Omega}_K(t)$ varies linearly over the interval $t \in [0,h]$ from an initial value $\widehat{\Omega}_1$ to a final value $\widehat{\Omega}_2$:
\begin{equation}
    \widehat{\Omega}_K(t) = \widehat{\Omega}_1 + \frac{t}{h}\left(\widehat{\Omega}_2 - \widehat{\Omega}_1\right)
\end{equation}
the integrals in \eqref{Magnus_expansion} evaluate analytically. The first-order term $\widehat{\Theta}_1(h)$ and the second-order non-commutative correction term $\widehat{\Theta}_2(h)$ are given explicitly by:
\begin{align}
    \widehat{\Theta}_1(h) &= \frac{h}{4} \left( \widehat{\Omega}_1 + \widehat{\Omega}_2 \right) \\[1em]
    \widehat{\Theta}_2(h) &= \frac{h^2}{48} \left( \widehat{\Omega}_2\widehat{\Omega}_1 - \widehat{\Omega}_1\widehat{\Omega}_2 \right) = \frac{h^2}{48} \left[\widehat{\Omega}_2, \widehat{\Omega}_1\right]
\end{align}
Truncating the series at second order $m=2$, the cumulative matrix exponential parameter $\mathbf{\Theta}^{(2)}(h) = \widehat{\Theta}_1(h) + \widehat{\Theta}_2(h)$ utilized to advance the unit quaternion state via $\mathbf{q}(h) = \exp\left(\mathbf{\Theta}^{(2)}(h)\right)\mathbf{q}(0)$ yields the closed-form expression:
\begin{equation}
    \mathbf{\Theta}^{(2)}(h) = \frac{h}{4} \left( \widehat{\Omega}_1 + \widehat{\Omega}_2 \right) + \frac{h^2}{48} \left( \widehat{\Omega}_2\widehat{\Omega}_1 - \widehat{\Omega}_1\widehat{\Omega}_2 \right)
\end{equation}

We have found that a second-order Magnus expansion ($m=2$) is 
sufficient for our application, though higher-order methods could be readily 
employed. Notably, introducing higher-order terms into the Magnus expansion 
only marginally increases the computational overhead of the quaternion 
mapping itself, without drastically impacting the total execution cost of 
the broader estimation algorithm.
\section{The algorithm for satellite self-diagnostic: a single trajectory case}
\subsection{Optimization Algorithm using Exact Mapping with Elliptic Functions}
\label{sec:exact_elliptic}


We now formulate the attitude-only inertia-estimation problem for a torque-free satellite. Unlike methods that require known applied torques, the proposed approach uses passive rotational motion and estimates the normalized inertia tensor from attitude measurements alone.

Consider a freely rotating satellite for which attitude measurements are available but angular-velocity measurements are not. This situation may arise when gyroscopes are unavailable, unreliable, or absent, as in some non-cooperative observation scenarios.

\begin{figure}[ht]
    \centering
    \includegraphics[width=0.7\linewidth]{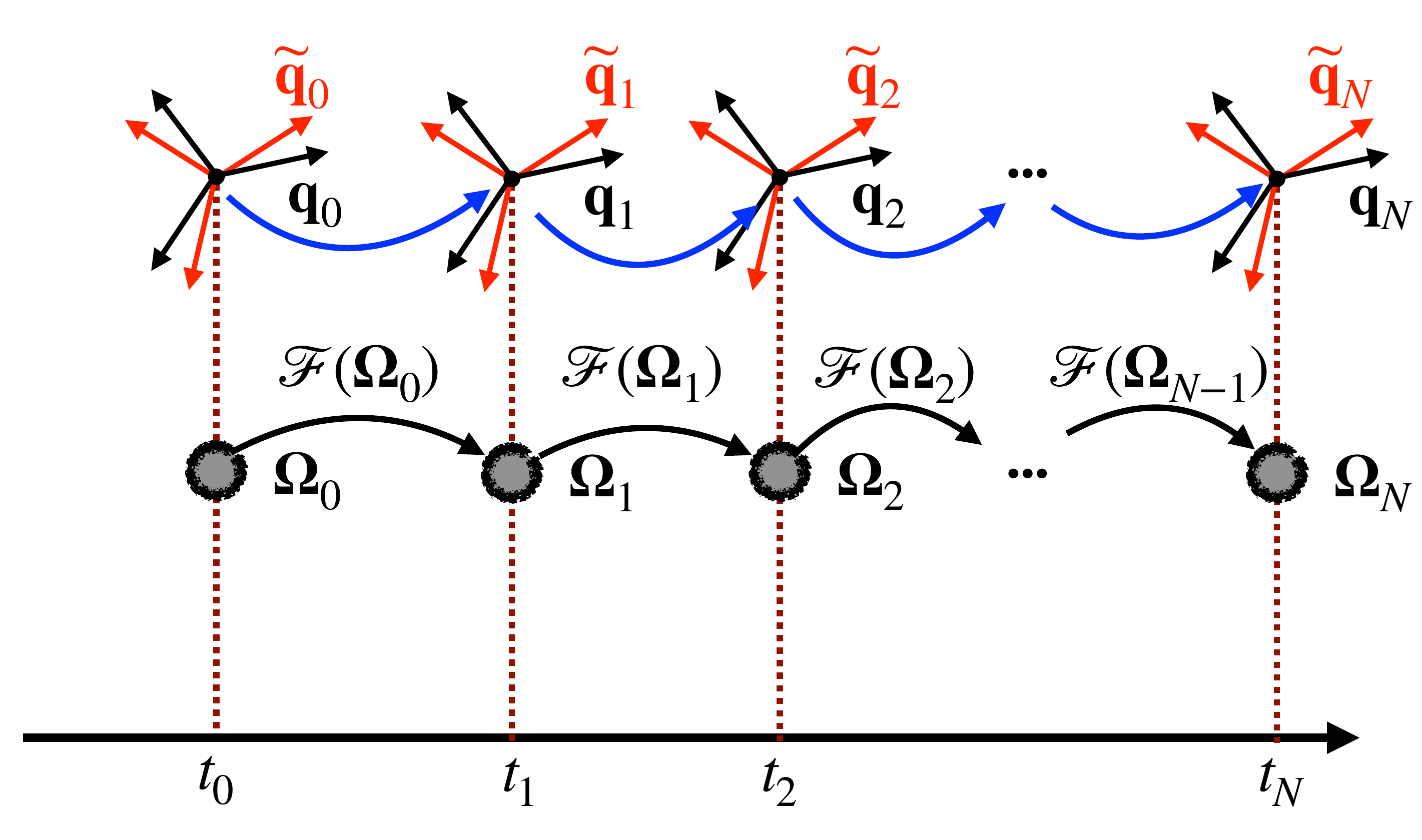}
    \caption{A schematics of the elliptic mapping method. The angular velocities are mapped using the exact elliptic mapping \eqref{omega_exact_map}, and the mapping between the quaternions is obtained using the Magnus expansion \eqref{Magnus_expansion}. }
    \label{fig:schematics}
\end{figure}

The idea for our method is illustrated on Figure~\ref{fig:schematics}. Suppose we know the sequence of attitude measurements $\widetilde{\mathbf{q}}(t_j)$, $j=0, \ldots, N$, but not the angular velocities measurement $\boldsymbol{\Omega}(t_j)$. Given initial conditions $(\boldsymbol{\Omega}_0,\mathbf{q}_0)$ and and a guess of the tensor of inertia $\mathbb{I}$, we can produce a trajectory $\boldsymbol{\Omega}_{j+1} = \mathcal{F}(\boldsymbol{\Omega}_j,\mathbb{I},h)$ using \eqref{omega_exact_map}, where $h$ is the time step, assumed the same for all data for simplicity. This sequence for angular velocity is then used to construct the sequence of quaternions $(\mathbf{q}_1,\ldots,\mathbf{q}_N)$ given the initial quaternion $\mathbf{q}_0$. 

The sequence of quaternions obtained from the simulated trajectory $\mathbf{q}_j$ is compared to the measured telemetry quaternions $\widetilde{\mathbf{q}}_j$ by computing the norm-squared of the vector part of the error quaternion: 
\begin{equation}
    \mathcal{L}(\boldsymbol{\Omega}_0,\mathbf{q}_0,\mathbb{I}) = \sum_{j=1}^N \left\| \left( \mathbf{q}_j^{-1} \otimes \widetilde{\mathbf{q}}_j\right)_{\rm vec} \right\|^2 \, 
    \label{loss_def}
\end{equation}
where $\mathbf{Q}_{\rm vec}$ denotes the vector component of a unit quaternion $\mathbf{Q} =(Q_0,Q_1,Q_2,Q_3)$, \emph{i.e.}, the three-dimensional vector $\mathbf{Q}_{\rm vec}=(Q_1,Q_2,Q_3)  \in \mathbb{R}^3$. The optimization procedure minimizes the objective function $\mathcal{L}(\boldsymbol{\Omega}_0,\mathbf{q}_0,\mathbb{I})$ defined by \eqref{loss_def}. Note that for small orientation errors, the vector part satisfies $\mathbf{Q}_{\rm vec} \approx \frac{1}{2}\boldsymbol{\theta}$, where $\boldsymbol{\theta} \in \mathbb{R}^3$ is the physical rotation vector. Minimizing \eqref{loss_def} is therefore dynamically equivalent to minimizing the true angular error vector magnitude across the trajectory.
Since we employ a nonlinear optimization procedure, it is imperative that we use reasonable initial guesses for the initial kinematic states $(\boldsymbol{\Omega}_0, \mathbf{q}_0)$ and the inertia matrix $\mathbb{I}$. To obtain this estimate, we utilize the Karush-Kuhn-Tucker (KKT) solution method, which leverages a linear formulation to provide a fast initial approximation. We found this warm start to be universally usable for our system across all simulated test cases. 

\subsection{KKT Estimate of Tensor of Inertia for a Single Trajectory}
\label{sec:KKT}

The nonlinear elliptic-shooting refinement requires a reliable initial estimate of the inertia tensor. We obtain this estimate from a linear least-squares formulation of the torque-free Euler equations. 

Let the symmetric inertia tensor be parameterized by a vector $\mathbf{Y}=[I_{xx}, I_{yy}, I_{zz}, I_{xy}, I_{xz}, I_{yz}]^T$.
At each time step $t_k$, the products $\mathbb{I}\dot{\boldsymbol{\Omega}}_k$ and $\mathbb{I}\boldsymbol{\Omega}_k$ can be written as linear functions of $\mathbf{Y}$:
\begin{equation}
\mathbb{I}\dot{\boldsymbol{\Omega}}_k
=
\mathbb{M}_d(\dot{\boldsymbol{\Omega}}_k)\mathbf{Y},
\qquad
\mathbb{I}\boldsymbol{\Omega}_k
=
\mathbb{M}_n(\boldsymbol{\Omega}_k)\mathbf{Y},
\end{equation}
where $\mathbb{M}_d,\mathbb{M}_n\in\mathbb{R}^{3\times 6}$. Substitution into the torque-free Euler equation gives a homogeneous linear constraint
\begin{equation}
\left[
\mathbb{M}_d(\dot{\boldsymbol{\Omega}}_k)
+
\boldsymbol{\Omega}_k^\times
\mathbb{M}_n(\boldsymbol{\Omega}_k)
\right]\mathbf{Y}
:=
\mathbb{A}_k\mathbf{Y}
=
\mathbf{0}.
\label{A_def}
\end{equation}
Stacking all time steps yields an over-determined linear system $\mathbb{A}\mathbf{Y}=\mathbf{0}$ for the unknown vector $\mathbf{Y}$.

Because the torque-free dynamics are scale invariant, the homogeneous system determines only the normalized inertia tensor. To remove the trivial solution and fix the scale, we impose
\begin{equation}
\mathbf{C}\mathbf{Y}=1,
\quad
\mathbf{C}=[1,1,1,0,0,0].
\end{equation}
Using the particular feasible vector $\mathbf{Y}_0=[1/3,1/3,1/3,0,0,0]^T$ and a basis $\mathbb{Q}\in\mathbb{R}^{6\times 5}$ for the null space of $\mathbf{C}$, the constrained variable is written as
\begin{equation}
\mathbf{Y}
=
\mathbf{Y}_0+\mathbb{Q}\mathbf{z},
\qquad
\mathbf{z}\in\mathbb{R}^5,
\label{z_def}
\end{equation}
with $\mathbf{C}\mathbb{Q}=0$. Substituting into the stacked system $\mathbb{A}\mathbf{Y}=\mathbf{0}$ gives the unconstrained least-squares problem
\begin{equation}
(\mathbb{A}\mathbb{Q})\mathbf{z}
=
-\mathbb{A}\mathbf{Y}_0 .
\end{equation}
Solving this system gives $\mathbf{z}$, and the initial inertia estimate is recovered from~\eqref{z_def}. 

This KKT-type constrained least-squares estimate is used only as a warm start for the nonlinear elliptic-shooting refinement. It is not the final estimator. As shown in the numerical results, subsequent refinement substantially improves accuracy relative to the linear initialization.

\subsection{Data Generation and Preprocessing}
\label{sec:data_generation}

Synthetic attitude datasets were generated to benchmark the estimation pipeline under telemetry conditions representative of the MYRIADE micro-satellite platform \cite{nainer2018orbit} and realistic onboard astronavigation accuracy.

\paragraph{Ground Truth Trajectory Generation}
A normalized true inertia tensor matching the physical properties of the MYRIADE platform was used: 
\begin{equation}
\mathbb{I}_{\text{true}} = \begin{bmatrix} 
31.3819 & -1.1136 & -0.2601 \\ 
-1.1136 &  21.1878 & -0.7783 \\ 
-0.2601 & -0.7783 &  35.7042 
\end{bmatrix},  \quad 
\mathbb{I}_{\text{norm}} 
= \begin{bmatrix} 
 0.3555 & -0.0126 & -0.0029 \\ 
-0.0126 &  0.2400 & -0.0088 \\ 
-0.0029 & -0.0088 &  0.4045 
\end{bmatrix}, 
\end{equation}
where, as we discussed, the normalized inertia matrix is defined so $\operatorname{tr}  \mathbb{I}_{\text{norm}}=1$, \emph{i.e.},  $\mathbb{I}_{\text{norm}}  = \mathbb{I}_{\text{true}}/\operatorname{tr}  (\mathbb{I}_{\text{true}})$
The system was initialized with a baseline angular velocity of magnitude  5 deg/s with the components projecting equally along the body axes, \emph{i.e.}, $\boldsymbol{\Omega}_0 = \frac{5}{\sqrt{3}}[1, 1, 1]^T  \, \text{deg/s}$. The ground truth rotational dynamics were simulated over a continuous $T=500$-second window with a discrete sampling interval of $h = 1.0$\,s using a high-order DOP853 explicit Runge-Kutta integrator with $10^{-13}$ accuracy, yielding $N = 501$ data points per trajectory. 

\paragraph{Noise Injection}

Attitude errors representative of star tracker measurements were simulated by perturbing each true quaternion with a random rotation. The perturbation axis was sampled uniformly in three dimensions, and the error angle was drawn uniformly from 0 to 30 arcseconds. This upper bound is conservative relative to legacy high-accuracy astronavigation systems \cite{liebe2002accuracy}, while remaining representative of commercial star trackers such as the Rocket Lab ST-RT, which reports attitude accuracy in the range of 5 to 55 arcseconds and update rates up to $5~\mathrm{Hz}$ \cite{rocketlab_startrackers}. The noisy attitude measurement was generated as
\begin{equation}
\widetilde{\mathbf{q}}_{k}
=
\mathbf{q}_k \otimes \mathbf{q}_{\text{noise},k},
\end{equation}
where $\otimes$ denotes quaternion multiplication.

\paragraph{Preprocessing and Smoothing}



The KKT initialization requires estimates of both angular velocity and angular acceleration. The noisy quaternion sequence was first unwrapped to avoid hemisphere jumps by enforcing $\mathbf{q} _ k\cdot\mathbf{q}_{k-1} \geq0$. Angular velocity was then approximated from successive attitude increments using the logarithmic map:
\begin{equation}
\boldsymbol{\Omega}_{k}
=
\frac{\operatorname{rotvec}(\mathbf{q}_{k-1}^{-1}\otimes\mathbf{q}_k)}{h}.
\end{equation}
To reduce differentiation-induced noise, the angular-velocity sequence was smoothed with a Savitzky-Golay filter using a 51-sample window and a third-order polynomial. The same filter configuration was used to compute the corresponding first derivative, providing the angular acceleration $\dot{\boldsymbol{\Omega}}$ used to construct the KKT matrices in Eq.~\eqref{A_def}.

\subsection{Simulation results and comparision with other methods} 
The initial guess from the KKT method is then fed to the optimization procedure described in Section~\ref{sec:exact_elliptic}. In order to provide a rigorous benchmark for our method, we also implement an Extended Kalman Filter (EKF) initialized with the same conditions obtained from the KKT warm start. While a precise initial guess may not be strictly necessary for simple linear systems, a robust ``warm start'' is crucial for the stability and convergence of an EKF applied to highly nonlinear satellite telemetry. The optimization procedure solves for the initial conditions $(\boldsymbol{\Omega}_0, \mathbf{q}_0)$ and the reduced vector $\mathbf{z} \in \mathbb{R}^5$ encoding the inertia matrix according to \eqref{z_def}, thereby automatically enforcing $\operatorname{tr}(\mathbb{I}) = 1$ across all evaluated solutions. 

In Table~\ref{tab:mc_comparison}, we present a performance comparison of our proposed method against the standard state-of-the-art EKF. Our optimization method improves upon the raw KKT initial results by two orders of magnitude, whereas the EKF yields an improvement of only one order of magnitude. Furthermore, our method is approximately 100 times more computationally efficient than the EKF, making it an ideal candidate for resource-constrained on-board processing. The benchmark results presented in Table~\ref{tab:mc_comparison} were evaluated over $N=100$ Monte Carlo simulation runs for both approaches.
\begin{table}[htbp]
\caption{Performance comparison of KKT, KKT+EKF, and KKT+Elliptic over $N=100$ realizations.}
\label{tab:mc_comparison}
\centering
\begin{tabular}{lcc}
\hline 
\hline
\textbf{Method} & \textbf{Mean Abs Error} & \textbf{Computational Time (s)} \\
\hline
KKT & 2.65e-05 & 0.0027 \\
KKT + EKF & 2.27e-06 & 11.1043 \\
KKT + Elliptic & 3.56e-07 & 0.1849 \\
\hline
\end{tabular}
\end{table}

\begin{figure}
    \centering
    \includegraphics[width=0.6\linewidth]{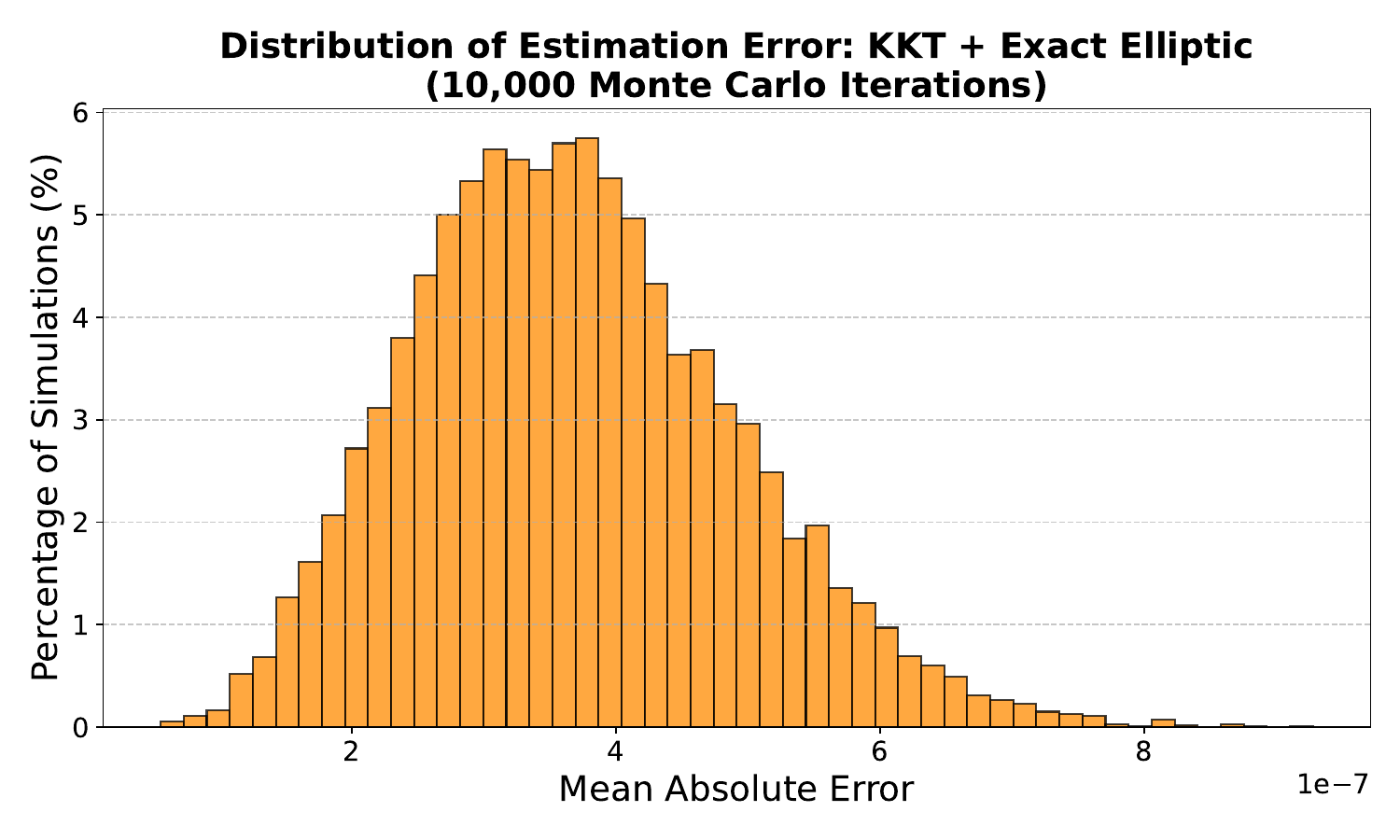}
    \caption{Histogram of inertia-estimation MAE over 10,000 Monte Carlo realizations for a single trajectory with $\boldsymbol{\Omega}_0=(1,1,1)^T5/\sqrt{3}~\mathrm{deg/s}$. Attitude samples are perturbed by random rotations with angles uniformly distributed over $[0,30]$ arcseconds, and KKT estimates initialize the elliptic refinement.}
    \label{fig:histogram}
\end{figure}

In Figure~\ref{fig:histogram}, we present the Mean Absolute Error (MAE) distribution obtained from a 10,000-run Monte Carlo simulation of our method under independent noise realizations. The dataset was generated from a single true trajectory subjected to a uniform $5$~arcsecond orientation tracking noise as described above. As illustrated, the resulting estimation error for the normalized, dimensionless inertia tensor is bounded between $1 \times 10^{-7}$ and $8 \times 10^{-7}$. 

\section{Observability studies}
\label{sec:observability}

\subsection{Global Observability Analysis}

We next examine how the initial angular-velocity direction affects the observability of the normalized inertia tensor. For torque-free motion, each trajectory lies on a polhode determined by the kinetic energy and angular momentum invariants. Pure rotation about a principal axis provides no usable information beyond identifying that axis as principal, while small deviations from the stable major or minor axes produce low-amplitude precessional motion that becomes difficult to distinguish from measurement noise. Near the unstable intermediate axis, finite-time observability can also degrade because the trajectory slows near the separatrix. The purpose of this analysis is to quantify these direction-dependent information losses.

Initial angular-velocity directions were sampled using a Fibonacci grid with $N=500$ points on the unit sphere. Each direction was assigned a fixed magnitude of $5^\circ/\mathrm{s}$, isolating the effect of direction from spin rate. For each initial condition, the ground-truth trajectory was generated by integrating the coupled Euler and quaternion kinematics in Eq.~\eqref{ext_Euler_eq} using the DOP853 integrator with relative and absolute tolerances of $10^{-13}$. Each trajectory was propagated for $1000~\mathrm{s}$ with a $1~\mathrm{s}$ sampling interval.

For each grid point, 10 independent Monte Carlo realizations were generated using the same attitude-noise and preprocessing procedure described in Section~\ref{sec:data_generation}. The initial inertia estimate was then computed using the KKT constrained least-squares method in Section~\ref{sec:KKT}, where the smoothed angular velocity and acceleration histories are assembled into the homogeneous linear system $\mathbf{A}\mathbf{I}=\mathbf{0}$.

\paragraph{Separatrix Proximty Handling:} To prevent gradient divergence near the unstable intermediate axis $I_2$, the algorithm dynamically tracked the dimensionless distance to the separatrix $\left| L^2 - 2TI_2 \right| / L^2$. If the trajectory fell into the critical zone (distance $< 10^{-2}$), the pipeline automatically switched from the analytical solver to the smooth numerical integrator DOP853, which completely eliminated estimation artifacts.

\subsection{Estimation Results}
\begin{figure}[htbp]
    \centering
    \includegraphics[width=0.7\textwidth]{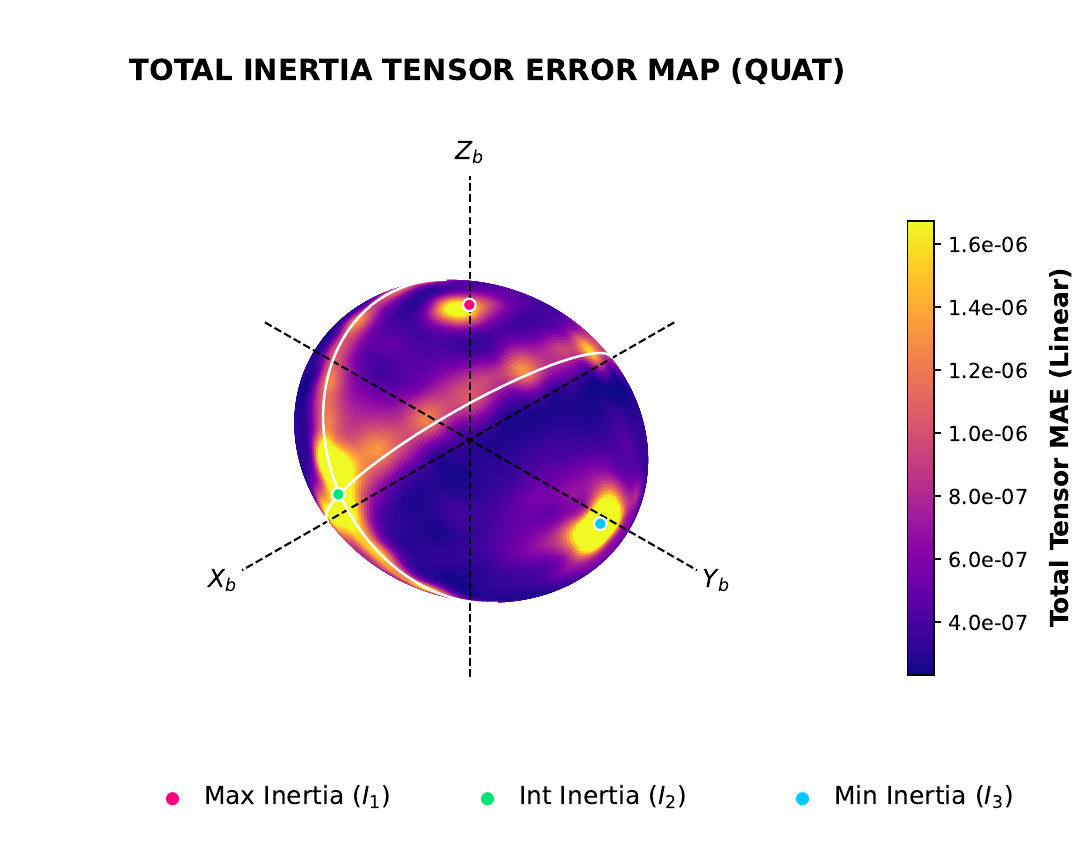}
    \caption{Global error map of the inertia tensor estimation (Total Tensor MAE). The separatrices (white lines) separate zones of observability, while local error maxima reveal the physical precessional degeneracy near the axes of maximum ($I_1$) and minimum ($I_3$) inertia.}
    \label{fig:inertia_error_map}
\end{figure}

The results of our simulations are presented as a global map of the mean absolute error (MAE) of the inertia tensor (see Fig.~\ref{fig:inertia_error_map}). One can see that while the accuracy is very good over most of the energy ellipsoid, it has worsening of accuracy close the the inertia axis (red, green and blue dots) and separatracies, marked as white curves. 

We now investigate the observability singularities along the principal axes of inertia in greater detail. As noted previously, pure rotation about an exact principal inertia axis yields no information regarding the structural components of the inertia tensor. This naturally raises the question of how observability degrades as the system's trajectory approaches these degenerate rotational states. As demonstrated below, the decay of observability in the neighborhood of principal axis rotations can be quantified utilizing concepts from the asymptotic solutions about the stable and unstable axes.


\subsection{Observability  Limits Near Principal-Axis Rotations}

To understand the fundamental limits of inertia tensor identification using our method, we analyze the structural tracking errors through the lens of local sensitivity. Let \(\boldsymbol{\theta}=\mathbf z\) denote the minimal parameterization of the normalized inertia tensor elements to be identified, as defined in \eqref{z_def}. We use the notation
\[
\boldsymbol{\theta}=(I_{12},I_{13},I_{22},I_{23},I_{33}).
\]
We separately consider the behavior near the stable principal axes, corresponding to the largest and smallest inertia eigenvalues, and near the unstable intermediate principal axis.

\subsubsection{Stable Principal Axes}

We first consider how small attitude-measurement errors are amplified as the transverse motion about a stable spin axis becomes small. Assume, without loss of generality, that the nominal stable spin axis is the first principal axis \(\mathbf Q_1\), namely the eigenvector of \(\mathbb I\) associated with the principal moment \(I_1\), \emph{i.e.}, \(\mathbb I\mathbf Q_1=I_1\mathbf Q_1\). For the exact spin about the first axis, \(\bOmega=\Omega_0\mathbf Q_1\), Euler's equations \eqref{ext_Euler_eq} are satisfied identically for any constant \(\Omega_0\), since \(\bOmega\times\mathbb I\bOmega=\Omega_0^2 I_1\,\mathbf Q_1\times\mathbf Q_1=0\). Thus, the orientation of the active principal axis is determined at leading order by the exact spin condition itself.

The degeneracy arises from the transverse directions that oscillate about the axes \((\mathbf Q_2,\mathbf Q_3)\). Suppose the main rotation is about the \(\mathbf Q_1\) axis and write
\[
\bOmega(t)
=
\Omega_0\mathbf Q_1
+
\epsilon V(t)\mathbf Q_2
+
\epsilon W(t)\mathbf Q_3
+
O(\epsilon^2).
\]
The leading transverse dynamics are then given by
\begin{equation}
\left\{
\begin{aligned}
\dot V
&=
-\Omega_0\frac{I_1-I_3}{I_2}\,W,
\\[1mm]
\dot W
&=
\Omega_0\frac{I_1-I_2}{I_3}\,V,
\end{aligned}
\right.
\quad\Rightarrow\quad
\begin{pmatrix}
V(t)\\[2mm]
W(t)
\end{pmatrix}
=
\begin{pmatrix}
\cos(\Omega t)
&
-\sqrt{
\frac{(I_1-I_3)I_3}
{(I_1-I_2)I_2}
}
\sin(\Omega t)
\\[4mm]
\sqrt{
\frac{(I_1-I_2)I_2}
{(I_1-I_3)I_3}
}
\sin(\Omega t)
&
\cos(\Omega t)
\end{pmatrix}
\begin{pmatrix}
V(0)\\[2mm]
W(0)
\end{pmatrix}.
\end{equation}
Here the transverse oscillation frequency is
\[
\Omega
=
\Omega_0
\sqrt{
\frac{(I_1-I_2)(I_1-I_3)}
{I_2I_3}
}.
\]
The physical angular-velocity component transverse to the nominal spin axis, 
\begin{align}
    \boldsymbol{\Omega}_{\perp}(t) =
 \epsilon V(t)\mathbf Q_2+\epsilon W(t)\mathbf Q_3,
\end{align}
is $O(\epsilon)$. Since the trajectories $\boldsymbol{\Omega(t)}$ are closed and the system is integrable,  the corresponding attitude sensitivity to the transverse inertia structure is also $O(\epsilon)$ over a fixed observation interval.

Let \(\mathbf r(\boldsymbol{\theta})\) denote the stacked attitude residual vector used in the least-squares problem \eqref{loss_def}, with individual residuals \(\mathbf r_k(\boldsymbol{\theta})=\operatorname{vec}\left(q(t_k;\boldsymbol{\theta})^{-1}\otimes \widetilde q_k\right)\). Let \(\mathbb S_\epsilon^{(\boldsymbol{\theta})}=\frac{\partial \mathbf r}{\partial\boldsymbol{\theta}}\) be the corresponding local sensitivity matrix. 
In local parameter coordinates, we conceptually separate perturbations that change the orientations of the active principal axis from those that change the transverse inertia structure. Accordingly, the sensitivity matrix is partitioned as
\begin{align}
    \mathbb S_\epsilon^{(\boldsymbol{\theta})}=[S_{\mathrm{axis}}\; S_{\mathrm{trans}}], 
\end{align}
where $S_{\mathrm{axis}}$ and $S_{\mathrm{trans}}$ denote the sensitivities associated with these two types of perturbations, respectively. 
The leading-order spin condition gives \(S_{\mathrm{axis}}=O(1)\) and \(S_{\mathrm{trans}}=O(\epsilon)\). Therefore the Gauss--Newton curvature $\mathbb{K}$, that is, the second derivative of the squared residual with respect to \(\boldsymbol{\theta}\), has the block scaling
\begin{equation}
\mathbb{K}_\epsilon^{(\boldsymbol{\theta})} = \left(\mathbb S_\epsilon^{(\boldsymbol{\theta})}\right)^T
\mathbb S_\epsilon^{(\boldsymbol{\theta})}
=
\begin{pmatrix}
O(1) & O(\epsilon)\\
O(\epsilon) & O(\epsilon^2)
\end{pmatrix}.
\label{Curvature_matr_eig}
\end{equation}
From \eqref{Curvature_matr_eig}, one can prove that in the absence of additional degeneracies, the eigenvalues of the square matrix $\mathbb{K}_\epsilon^{(\boldsymbol{\theta})}$ are $O(\epsilon^2)$. 
Consequently, from \eqref{Curvature_matr_eig}, the singular values of $S_\epsilon^{(\boldsymbol{\theta})}$ are square roots of eigenvalues of $\mathbb{K}_\epsilon^{(\boldsymbol{\theta})}$, and  we thus have
\begin{equation}
\sigma_{\min}(S_{\mathrm{trans}})
=
c\epsilon+O(\epsilon^2),
\qquad c>0 .
\label{eq_min_eig}
\end{equation} 
For a small perturbation \(\boldsymbol{\eta}\) in the residual space, the first-order residual perturbation satisfies
\[
\delta\mathbf r
\simeq
\mathbb S_\epsilon^{(\boldsymbol{\theta})}\delta\boldsymbol{\theta}
+
\boldsymbol{\eta}.
\]
Restricting attention to the weak transverse directions gives
\[
S_{\mathrm{trans}}\delta\boldsymbol{\theta}_{\mathrm{trans}}
\approx
-\boldsymbol{\eta}.
\]
Thus
\[
\|\delta\boldsymbol{\theta}_{\mathrm{trans}}\|
\lesssim
\frac{\|\boldsymbol{\eta}\|}
{\sigma_{\min}(S_{\mathrm{trans}})}
=
O(\epsilon^{-1})\|\boldsymbol{\eta}\|.
\]
Hence, for a fixed observation time, the transverse inertia directions lose conditioning as the motion approaches an exact stable principal-axis rotation. The curvature in these directions scales as \(O(\epsilon^2)\), while the corresponding parameter uncertainty scales as \(O(\epsilon^{-1})\).

This behavior is observed in the left panel of Figure~\ref{fig:observability_limits_combined}: as the sampled rotations approach either stable axis, the estimated uncertainty increases approximately as \(1/\epsilon\) over the neighborhood of the axis. Here, we define 
\(
\epsilon = \|\boldsymbol{\Omega}_{\perp}\|/\|\boldsymbol{\Omega}_{\parallel}\|
\)
as the transverse angular-velocity offset from the principal axis. 
\subsubsection{Unstable intermediate axis \texorpdfstring{(\(I_2\))}{(I2)}}

The dynamics near the intermediate principal axis \(I_2\) are locally
hyperbolic and are bounded by heteroclinic separatrices
\cite{arnol2013mathematical}.  As the initial condition approaches the
pure intermediate-axis rotation, nearby trajectories spend increasingly
long intervals near the hyperbolic equilibrium.  Consequently, the
characteristic period of the associated separatrix motion diverges
logarithmically:
\begin{equation}
    T_{\mathrm{sep}}
    =
    C|\ln \epsilon| + O(1),
    \qquad C>0 ,
    \label{T_sep_divergence}
\end{equation}
where \(\epsilon\) measures the initial distance from the intermediate
axis.

\begin{figure*}[tbp]
    \centering
    \includegraphics[width=\textwidth]{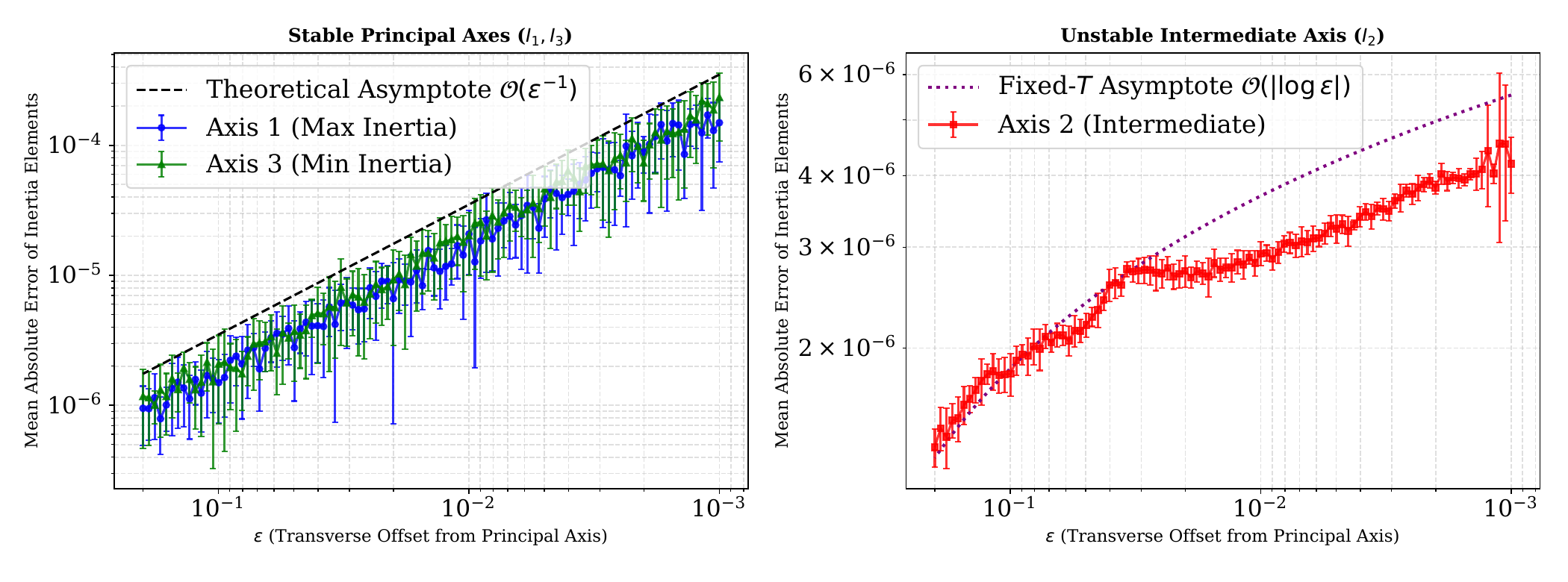}
    \caption{Inertia-tensor parameter errors versus transverse perturbation magnitude $\epsilon$. Stable-axis cases follow the $\mathcal{O}(\epsilon^{-1})$ degradation trend, while the intermediate-axis case follows a weaker finite-time logarithmic trend. Error bars show $\pm1\sigma$ over 10 noisy realizations.}
    \label{fig:observability_limits_combined}
\end{figure*}

Each separatrix excursion traverses an \(O(1)\) portion of the phase-space curve, but the time required for such an excursion grows progressively longer as \(\epsilon \to 0\).  For a fixed measurement time \(T\), the number of observable macroscopic separatrix excursions therefore decreases as
\begin{equation}
    N_{\mathrm{flip}}
    \sim
    \frac{T}{T_{\mathrm{sep}}}
    =
    O\!\left(\frac{1}{|\log \epsilon|}\right).
\end{equation}
Thus, over a fixed observation window, the accumulated phase-space traversal decreases as
$N_{\mathrm{flip}} \sim |\log \epsilon|^{-1}$.
Consequently, the intermediate-axis degradation is not caused by an algebraic loss of transverse signal, as in the stable-axis case.  It is instead governed by the logarithmic separatrix time
\(T_{\mathrm{sep}}\sim |\log\epsilon|\).  Hence any associated accuracy loss is expected to be logarithmic, or at most governed by the powers of \(|\log\epsilon|\), rather than by an algebraic power of \(1/\epsilon\).  This prediction is verified in the right panel of Figure~\ref{fig:observability_limits_combined}.

\color{black} 

We next evaluate whether the proposed inertia-estimation framework remains effective when the attitude inputs are obtained from image-based pose estimation rather than prescribed noise models.

\FloatBarrier
\section{Image-based attitude measurements using observations from a chaser spacecraft}
\subsection{Image Simulation Framework}
\subsubsection{Overview of the Application Pipeline}
\label{subsec:application_pipeline}
\begin{figure}[htbp]
    \centering
    \includegraphics[width=0.9\textwidth]{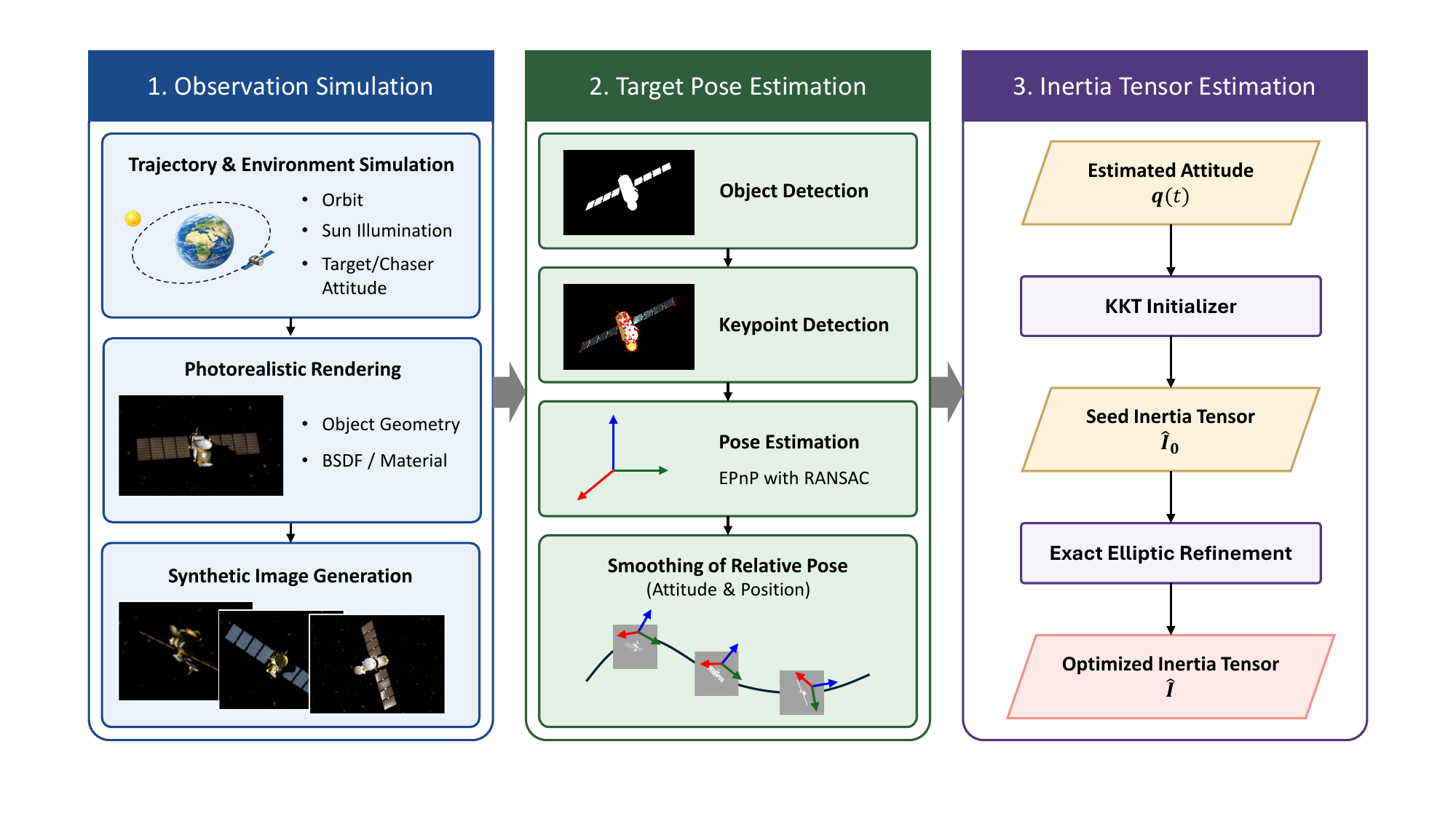}
    \caption{Flowchart of image-based simulation framework}
    \label{fig:overview_pipeline}
\end{figure}
This section replaces direct attitude measurements with estimates derived from synthetic spacecraft images. The objective is to test whether the estimator remains effective when its input is generated by a vision-based pose-estimation pipeline, where errors depend on target geometry, illumination, viewing direction, and keypoint-detection quality.

Figure~\ref{fig:overview_pipeline} summarizes the overall simulation and estimation workflow. The target and chaser motion are simulated in a simplified rendezvous and proximity operations (RPO) scenario in low Earth orbit (LEO). At each observation epoch, the relative pose, target attitude, camera model, and Sun direction are passed to Mitsuba~3~\cite{Mitsuba3} to render spacecraft images. The images are then processed by a vision pipeline that detects body-frame keypoints and solves a perspective-$n$-point problem to recover the target attitude relative to the chaser camera. The resulting attitude sequence is preprocesed and supplied to the inertia tensor estimator. 

This image-based evaluation introduces geometry- and illumination-dependent attitude errors caused by viewing direction, shadowing, partial visibility, keypoint uncertainty, and pose-recovery errors. These effects are not well represented by independent Gaussian attitude noise but arise naturally in optical observations of a non-cooperative spacecraft. 

\subsubsection{Relative Orbit, Attitude, and Illumination Simulation}
\label{subsec:relative_orbit_attitude_illumination}

The relative motion and attitude histories are generated using a simplified RPO scenario in LEO. The target follows a circular reference orbit, and the chaser motion is propagated in the target-centered Hill frame using the Clohessy-Wiltshire model. The corresponding mean motion is determined by the sampled target altitude, and the initial relative state is selected to produce a bounded inspection trajectory with representative close-proximity imaging ranges.

At each time step, the relative trajectory defines the line-of-sight vector from the chaser to the target, and the chaser camera is oriented toward the target. The Sun direction is propagated in the Earth-centered inertial frame and transformed into the Hill frame for rendering. These quantities define the viewing and illumination geometry used to generate each image. Challenging cases, including partial shadowing, low-contrast views, and backlit observations, are retained because they represent realistic error sources for the downstream pose-estimation pipeline.

The target attitude is propagated independently using the torque-free rigid-body equations, including the orientation dynamics in Eq.~\eqref{ext_Euler_eq}, with the reference inertia tensor described in Sec.~\ref{subsec:rendering}. The resulting attitude is transformed into the Hill frame before rendering. This transformation places the target attitude, chaser line-of-sight geometry, and Sun direction in a common frame for image generation.

\subsubsection{Photorealistic Image Rendering}
\label{subsec:rendering}

The JASON-1 spacecraft is selected as the target object because it represents a generic box-wing satellite architecture, comprising a central bus, solar panels, antennas, and external appendages~\cite{podaac_jason1}. The three-dimensional model is obtained from NASA's 3D Resources repository~\cite{nasa_jason1_3d} and scaled to match the physical dimensions of the spacecraft. The model is converted into a watertight mesh, and the reference inerti atensor is computed from the resulting closed triangular surface using the divergence-theorem-based method \cite{semechko2026rigid}, under the assumption of uniform density. The true inertia matrix and its normalized counterpart are  
\begin{align}
    \mathbb{I}_{\text{true}} = 
    \begin{bmatrix}
        2264.4 &  -16.5 &  -1.2 \\
        -16.5 &  1691.5 &   -0.1 \\
        -1.2 &   -0.1  & 850.9
    \end{bmatrix} \ \text{kg}\cdot\text{m}^2, \quad 
    \mathbb{I}_{\text{norm}} = 
    \begin{bmatrix}
        0.4711 & -0.0034 & -0.0003 \\
        -0.0034  & 0.3519 &  -0.0000 \\
        -0.0003 & -0.0000 &  0.1770
    \end{bmatrix}.
\end{align}

\begin{figure}
    \centering
    \includegraphics[width=0.6\linewidth]{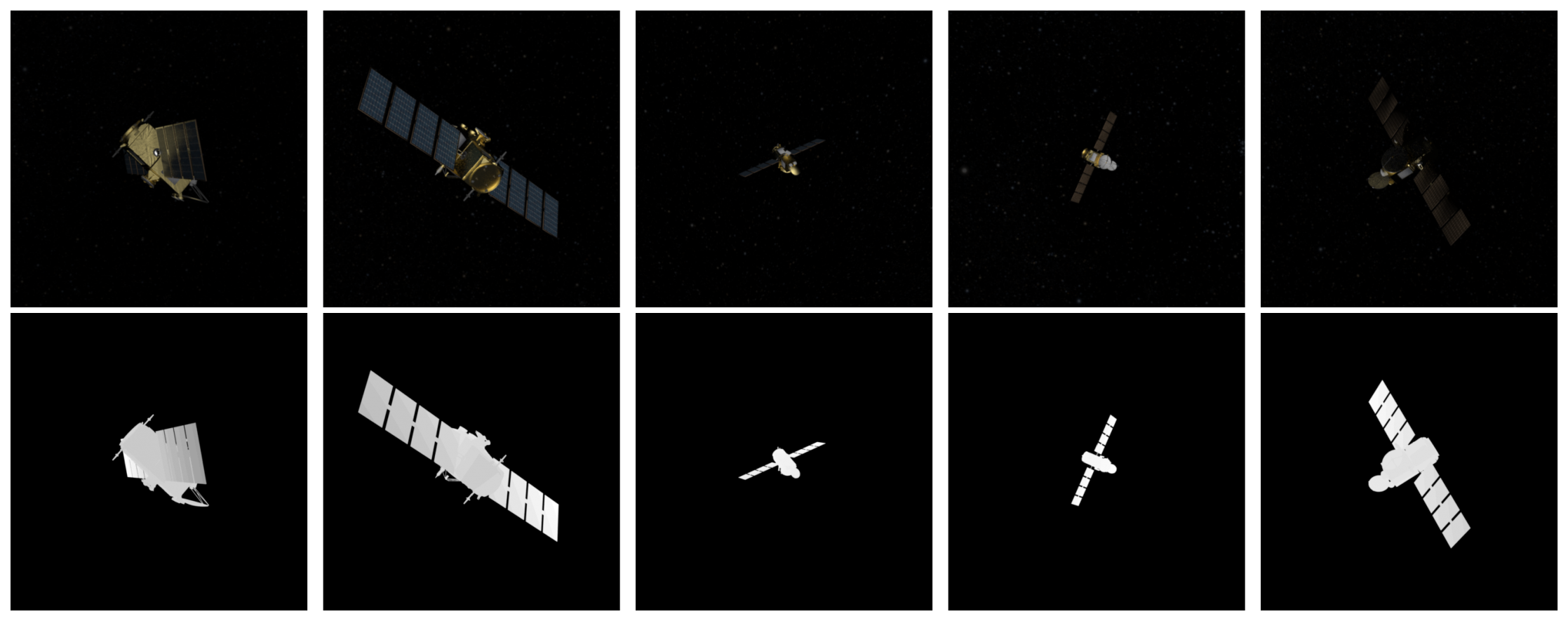}
    \caption{Example of rendered images with various range and lighitng conditions. Top: RGB images, bottom: binary mask images.}
    \label{fig:example_images_jason1}
\end{figure}

Target images are rendered using Mitsuba~3, a physically based rendering system~\cite{Mitsuba3}. The chaser camera is modeled as a perspective camera with a $30^\circ$ field of view, and all images are rendered at $512 \times 512$ pixels. For each observation epoch, the renderer is supplied with the target pose, camera pose, Sun direction, and camera parameters obtained from the orbital and attitude simulation. A binary object mask is also generated for training the object-detection component of the pose-estimation pipeline. Representative rendered images are shown in Fig.~\ref{fig:example_images_jason1}, illustrating the variation in apparent geometry, shadowing, and image contrast encountered by the vision pipeline.

\subsubsection{Dataset Definitions}
\label{subsec:datasets}

\begin{table}[t]
    \centering
    \renewcommand{\arraystretch}{1.1} 
    \caption{Summary of generated image datasets.}
    \label{tab:datasets}
    \begin{tabular}{m{2.8cm} m{7cm} m{2.5cm} m{2.5cm}}
    \hline 
    \hline
    \centering Dataset & \centering Purpose & \centering Cases & \centering\arraybackslash Images per case \\
    \hline
    Random-view & Pose-estimation training and validation & \centering 10,000  views & \centering\arraybackslash 1 \\
    \hline 
    Single-arc RPO & Inertia tensor estimation from continuous torque-free attitude motion & \centering 100 trajectories& \centering\arraybackslash 2,000 \\
    \hline 
    Multi-arc RPO & Inertia tensor estimation from stratified multi-arc attitude motion with two impulsive maneuvers & \centering 120 trajectories & \centering\arraybackslash 900 \\
    \hline
    \end{tabular}
\end{table}

Three image datasets are generated, as summarized in Table~\ref{tab:datasets}. The \textit{random-view dataset} is used to train and validate the vision-based pose-estimation pipeline, whereas the \textit{single-arc RPO} and \textit{multi-arc RPO} datasets are used to evaluate the proposed inertia tensor estimator under image-based attitude measurements.

The \textit{random-view dataset} consists of 10,000 independently rendered images with randomized camera-target geometry and illumination. The camera-target distance is sampled uniformly from $30$ to $100$~m, and the viewing direction is sampled uniformly over the sphere. For each view, the illumination direction is sampled within a $45^\circ$ cone about the viewing direction, limiting the phase angle to $90^\circ$. This retains challenging illumination conditions while avoiding fully dark target images. The dataset is split into training and validation subsets using an 8:2 ratio.

The \textit{single-arc RPO dataset} consists of 100 continuous torque-free RPO image sequences, each containing 2,000 images. The target altitude is sampled from $400$ to $600$~km, and the chaser initial relative state is selected so that the camera-target distance remains within $35$ to $100$~m. The target initial attitude is randomized, the initial angular-velocity direction is sampled uniformly on the unit sphere, and its magnitude is sampled from a Gaussian distribution with mean $0.01$~rad/s and standard deviation $0.001$~rad/s. The Sun direction is randomized subject to a maximum phase angle of $60^\circ$. This dataset is used to assess whether the observability trends predicted from ideal attitude measurements persist when the attitude sequence is obtained from rendered images.

The \textit{multi-arc RPO dataset} evaluates the estimator under discontinuous attitude dynamics. Each trajectory contains two impulsive angular-velocity changes, dividing the attitude history into three torque-free arcs. Images are generated at a fixed cadence of $\Delta t=1.0$~s, with $N=900$ images per trajectory and total duration $T=899$~s. This fixed-cadence design ensures that differences across trajectories are driven primarily by the prescribed spin regime and maneuver class rather than by temporal sampling.

The multi-arc dataset is generated using a $3\times4$ stratified sampling grid consisting of three initial spin regimes and four maneuver classes, as summarized in Table~\ref{tab:multi_arc_stratification}. The maneuver classes are denoted by SS, SL, LS, and LL, where each letter indicates whether the corresponding impulsive maneuver belongs to the small or large angular-velocity-increment bin. For example, SL denotes a trajectory with a small first maneuver and a large second maneuver.

\begin{table}[t]
\centering
\caption{Stratification of the multi-arc RPO dataset.}
\label{tab:multi_arc_stratification}
\begin{tabular}{lll}
\hline
\hline
\textbf{Category} & \textbf{Class} & \textbf{Range} \\
\hline
Initial spin & Low    & $\|\boldsymbol{\omega}_0\|\in[0.005,0.015]$ rad/s \\
& Medium & $\|\boldsymbol{\omega}_0\|\in[0.025,0.055]$ rad/s \\
& High   & $\|\boldsymbol{\omega}_0\|\in[0.075,0.120]$ rad/s \\
\hline
Maneuver increment & Small & $\|\Delta\boldsymbol{\omega}\|\in[0.003,0.012]$ rad/s \\
& Large & $\|\Delta\boldsymbol{\omega}\|\in[0.025,0.060]$ rad/s \\
\hline
Maneuver class & SS, SL, LS, LL & Letter order denotes first/second maneuver \\
\hline
\end{tabular}
\end{table}

For each grid cell, orbital and illumination parameters are generated using the same procedure as in the single-arc RPO dataset. Initial attitudes are randomized, and initial angular-speed magnitudes are sampled log-uniformly from the assigned spin bin. Maneuver directions and magnitudes are sampled according to the assigned maneuver class.

Candidate trajectories are rejected if the post-maneuver angular speed falls outside $[0.003,0.16]$~rad/s, if the maneuver-induced direction change falls outside $[2^\circ,140^\circ]$, or if $\|\Delta\boldsymbol{\omega}\|/\|\boldsymbol{\omega}_{\mathrm{before}}\|>10$. Accepted trajectories are rendered using the Mitsuba-based image-generation pipeline together with the corresponding orbital, attitude, camera, and illumination metadata. Ten samples are drawn from each grid cell, yielding $120$ multi-arc trajectories.

\subsection{Image-Based Pose Estimation}
\label{subsec:image-based-pose-estimation}
\begin{figure}[htbp]
    \centering

    \begin{subfigure}[t]{0.42\textwidth}
        \centering
        \includegraphics[height=0.17\textheight,keepaspectratio]{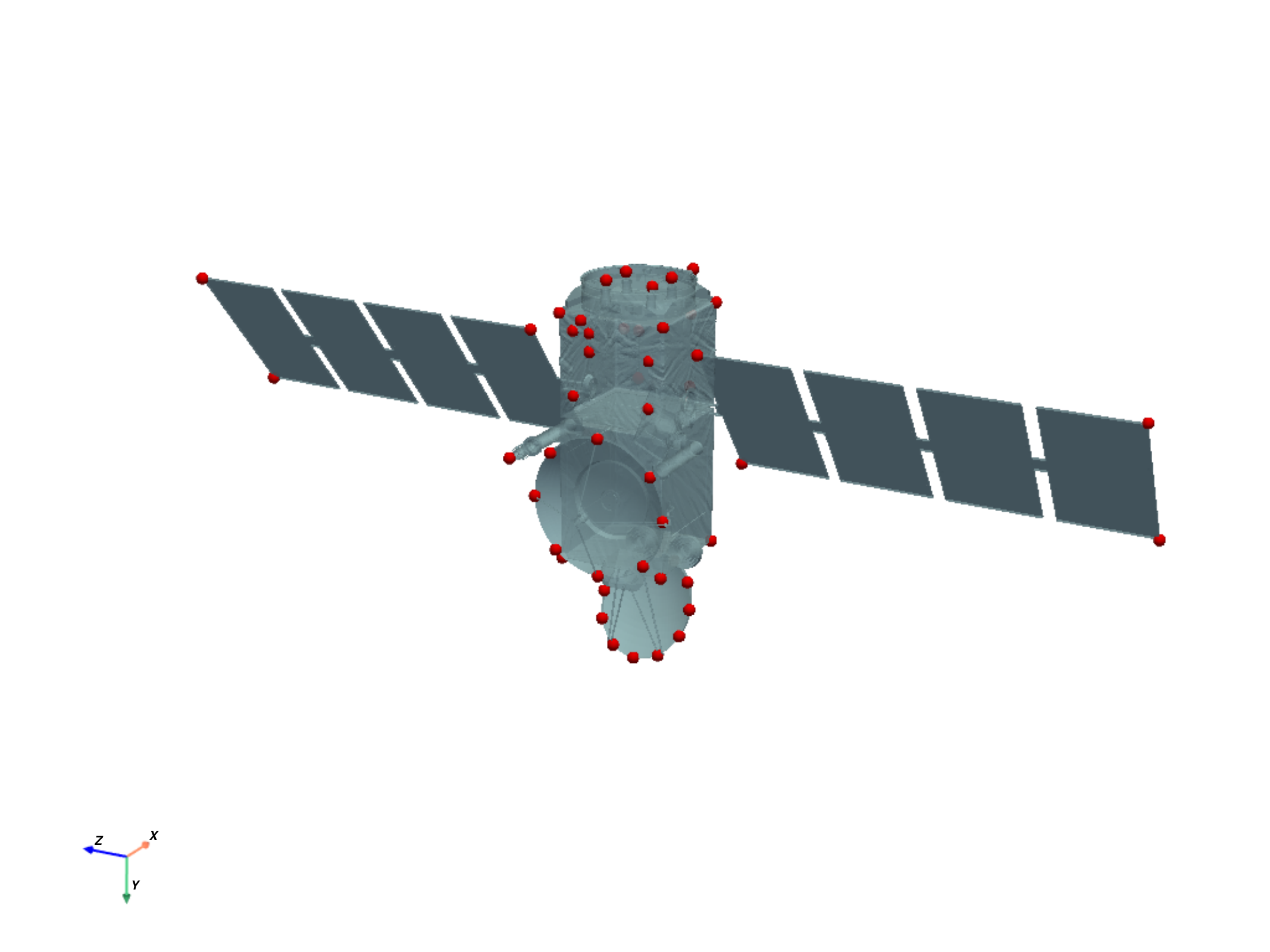}
        \caption{3D keypoint positions}
        \label{fig:keypoints_body_3d}
    \end{subfigure}
    \hfill
    \begin{subfigure}[t]{0.54\textwidth}
        \centering
        \includegraphics[height=0.17\textheight,keepaspectratio]{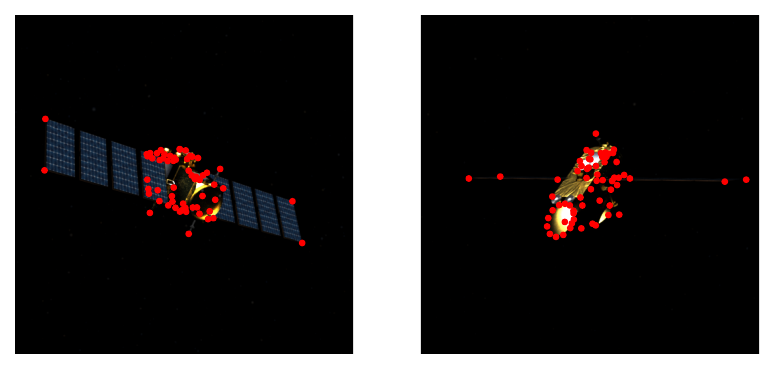}
        \caption{Examples of projected 2D keypoint positions}
        \label{fig:keypoints_body_projected}
    \end{subfigure}
    \caption{Definition and projection of spacecraft keypoints.}
    \label{fig:keypoint_body}
\end{figure}

The image-based pose estimation module follows the keypoint-based pipeline developed in our previous work \cite{kobayashi_conf_AMOS_2023}. Since the present study focuses on inertia estimation rather than pose-estimation algorithm development, only the main procedure is summarized here.

Given calibrated images, the relative pose is estimated from 2D--3D keypoint correspondences by solving the perspective-$n$-point problem,
\begin{align}
    \mathbf{u} \sim \mathbf{K}
    \renewcommand{\arraystretch}{0.6}
    \begin{bmatrix}
    \mathbf{R} & \mathbf{t}
    \end{bmatrix}
    \mathbf{X},
\end{align}
where $\mathbf{u}$ is the homogeneous image coordinate of a detected keypoint, $\mathbf{K}$ is the camera calibration matrix, $\mathbf{X}$ is the corresponding spacecraft body-frame keypoint, and $(\mathbf{R},\mathbf{t})$ is the transformation from the spacecraft frame to the camera frame.

The 3D keypoints are manually defined on visually distinguishable features of the JASON-1 model, including solar-panel corners, aperture edges, and bus corners, as shown in Fig.~\ref{fig:keypoints_body_3d}. The corresponding 2D keypoints are estimated using a two-stage vision pipeline. A YOLO-based detector \cite{jocher2026ultralyticsyolo26unifiedrealtime} first localizes the spacecraft bounding box, and an RTMPose-based keypoint detector \cite{jiang2023rtmposeunifiedrealtime} then estimates the keypoint locations within the detected region. Both networks are trained using the random-view dataset described in Sec.~\ref{subsec:datasets}.

After the 2D--3D correspondences are obtained, the relative pose is estimated independently for each frame using EPnP with RANSAC outlier rejection~\cite{lepetit_epnp_2009}. The resulting pose sequence is post-processed to suppress frame-to-frame jitter: outliers are detected using a Hampel filter and replaced by interpolation, with quaternion SLERP used for the attitude component. The smoothed attitude sequence is then supplied to the downstream pipeline for inertia tensor estimation.

\subsection{Pose Estimation Results}
\label{subsec:pose_estimation_results}

\begin{figure}[htbp]
    \centering
    \includegraphics[width=0.8\linewidth]{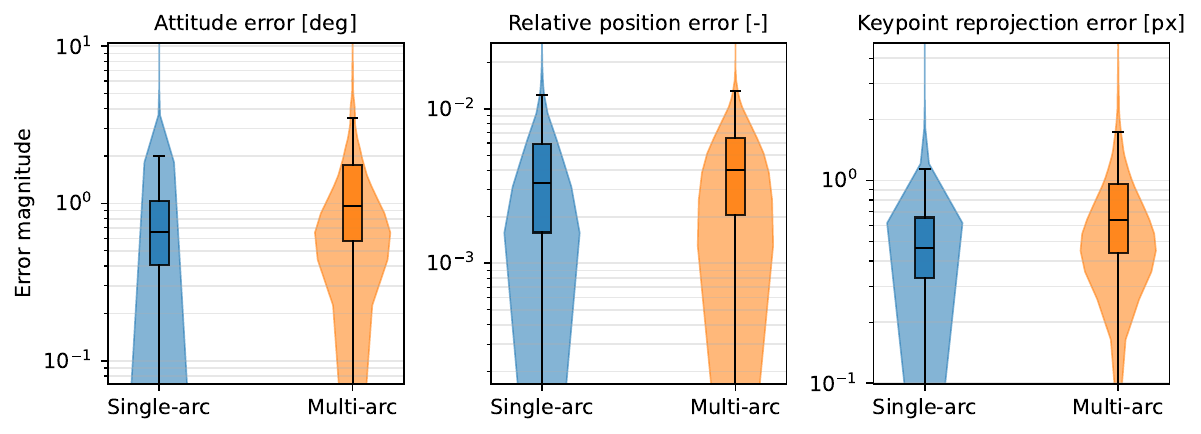}
    \caption{Pose estimation errors tested on single- and multi-arc RPO datasets.}
    \label{fig:pose_error_distributions}
\end{figure}

\begin{table}[htbp]
    \caption{Summary statistics of image-based pose-estimation errors for the single-arc and multi-arc RPO datasets.}
    \label{tab:pose_error_summary}
    \centering
    \renewcommand{\arraystretch}{1.1}
    \begin{tabular}{llrccc}
    \hline
    \hline
    \multicolumn{1}{c}{Metric} 
    & \multicolumn{1}{c}{Dataset} 
    & \multicolumn{1}{c}{Median} 
    & \multicolumn{1}{c}{90th perc.} 
    & \multicolumn{1}{c}{95th perc.} 
    & \multicolumn{1}{c}{99th perc.} \\
    \hline
    Attitude error [deg] 
        & Single-arc & 0.660 & 1.577 & 2.020 & 3.296 \\
        & Multi-arc  & 0.964 & 3.055 & 4.285 & 8.291 \\
    \hline
    Position error [\%] 
        & Single-arc & 0.330 & 0.889 & 1.130 & 1.812 \\
        & Multi-arc  & 0.399 & 0.909 & 1.092 & 1.543 \\
    \hline
    Reprojection error [px] 
        & Single-arc & 0.466 & 0.904 & 1.126 & 1.794 \\
        & Multi-arc  & 0.637 & 1.455 & 1.905 & 3.644 \\
    \hline
    \end{tabular}
\end{table}

The pose estimation pipeline is evaluated using the single-arc RPO and multi-arc RPO datasets. Three error metrics are used. The attitude error is defined as the geodesic distance between the predicted and true quaternions. The position error is defined as the relative error of the object-to-camera translation vector. The reprojection error is computed by projecting the known 3D keypoints onto the image plane using the estimated pose and then averaging their distances from the true image coordinates.
\begin{align}
    e_{\text{att}} = 2\cos^{-1}\left(\lvert \mathbf{q}_{\text{pred}}\cdot\mathbf{q}_{\text{true}}\rvert\right), 
    \quad
    e_{\text{pos}} = \frac{\|\mathbf{t}_{\text{pred}}-\mathbf{t}_{\text{true}}\|}{\|\mathbf{t}_{\text{true}}\|}, 
    \quad
    e_{\text{proj}} = \frac{1}{M}\sum_{i=1}^M \left\| \pi(K[\mathbf{R}_{\text{pred}} \ \mathbf{t}_{\text{pred}}]X_i) - u_{\text{true}, i} \right\|,
\end{align}
where M is the number of keypoints and $\pi(\cdot)$ denotes the perspective division from homogeneous image coordinates to pixel coordinates.

Figure~\ref{fig:pose_error_distributions} shows violin plots overlaid with box plots for the three pose-estimation error metrics on the single- and multi-arc RPO datasets. The corresponding summary statistics are reported in Table~\ref{tab:pose_error_summary}. Overall, the image-based pose-estimation pipeline achieves sub-degree median attitude accuracy for both datasets, with median attitude errors of $0.660^\circ$ and $0.964^\circ$ for the single- and multi-arc cases, respectively. The position error remains small in both cases, with median relative errors below 0.4\%. The reprojection error is also below one pixel at the median level for both datasets, indicating that the estimated poses are geometrically consistent with the projected keypoint observations.

The multi-arc dataset exhibits a broader error distribution than the single-arc dataset, particularly in attitude and reprojection error. The 99th-percentile attitude error increases from $3.296^\circ$ in the single-arc case to $8.291^\circ$ in the multi-arc case, and the 99th-percentile reprojection error increases from $1.794$ pixels to $3.644$ pixels. This degradation is expected because the multi-arc dataset contains more diverse rotational states and discontinuous attitude changes induced by impulsive maneuvers. In contrast, the position-error distributions are comparable between the two datasets, suggesting that the main difference between the datasets is in attitude estimation rather than translational pose estimation. These results indicate that the multi-arc dataset provides a more challenging test case for subsequent inertia tensor estimation, while still maintaining pose-estimation accuracy sufficient for image-based dynamical analysis.

\subsection{Inertia Tensor Estimation from Image-Derived Attitudes: A Single Trajectory Case}

The proposed inertia-estimation method is evaluated on the single-arc RPO dataset using the image-derived attitude histories from Section~\ref{subsec:pose_estimation_results}. Each sequence is generated from a fixed initial angular velocity and propagated under torque-free rotational motion. The attitude history is then reconstructed by the image-based pose-estimation pipeline and used as input to the inertia-estimation method.

Before estimation, frames with unreliable pose estimates are removed using a reprojection-error test. The known 3D keypoints are projected into the image using the estimated pose and compared with the keypoint locations predicted by the computer vision model. A frame is classified as an outlier if the median reprojection error exceeds 5 pixels. Outlier removal was required for 19 of the 100 datasets, with an average of 5.5 frames removed per dataset. The accuracy of the inertia estimate is evaluated using the mean absolute error of the tensor components.

\begin{figure}[ht]
    \centering
    \includegraphics[width=0.9\linewidth]{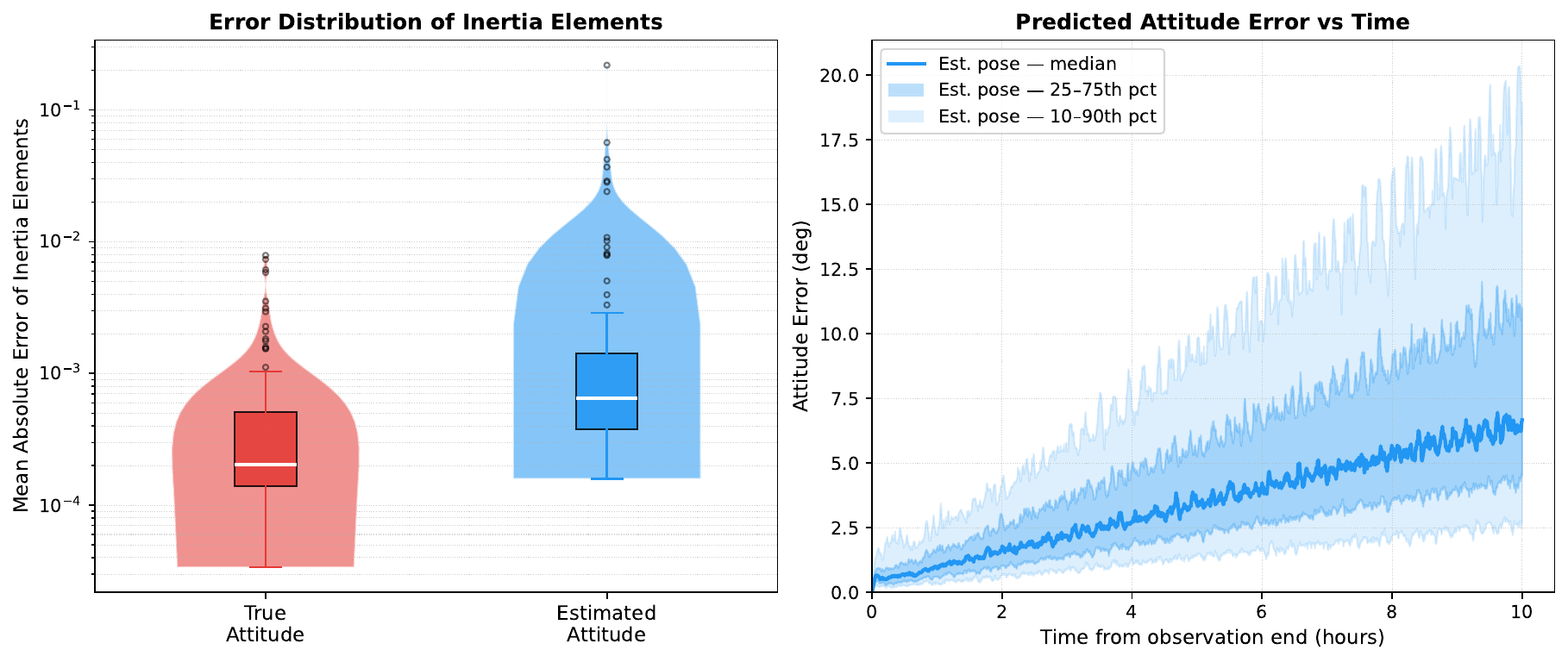}
    \caption{Single-arc RPO inertia-estimation performance. Left: distributions of inertia tensor MAE using true and image-based attitude inputs. Right: 10-hour attitude-prediction error obtained using the estimated inertia tensor.}
    \label{fig:results_single_arc_RPO}
\end{figure}

Figure~\ref{fig:results_single_arc_RPO} summarizes the single-arc RPO results and quantifies the degradation caused by replacing true attitudes with image-derived estimates. Using true attitudes, the median tensor MAE across the 100 datasets is $2.0\times10^{-4}$; using estimated attitudes, it increases to $6.5\times10^{-4}$. The 90th percentile similarly increases from $1.8\times10^{-3}$ to $8.3\times10^{-3}$. Thus, pose-estimation errors increase the inertia-estimation error, especially in less favorable cases. Nevertheless, the long-horizon prediction impact remains moderate: using the estimated inertia tensor, the median attitude prediction error after 10 hours is $6.7^\circ$.

\begin{figure}[htbp]
    \centering
    \includegraphics[width=0.9\linewidth]{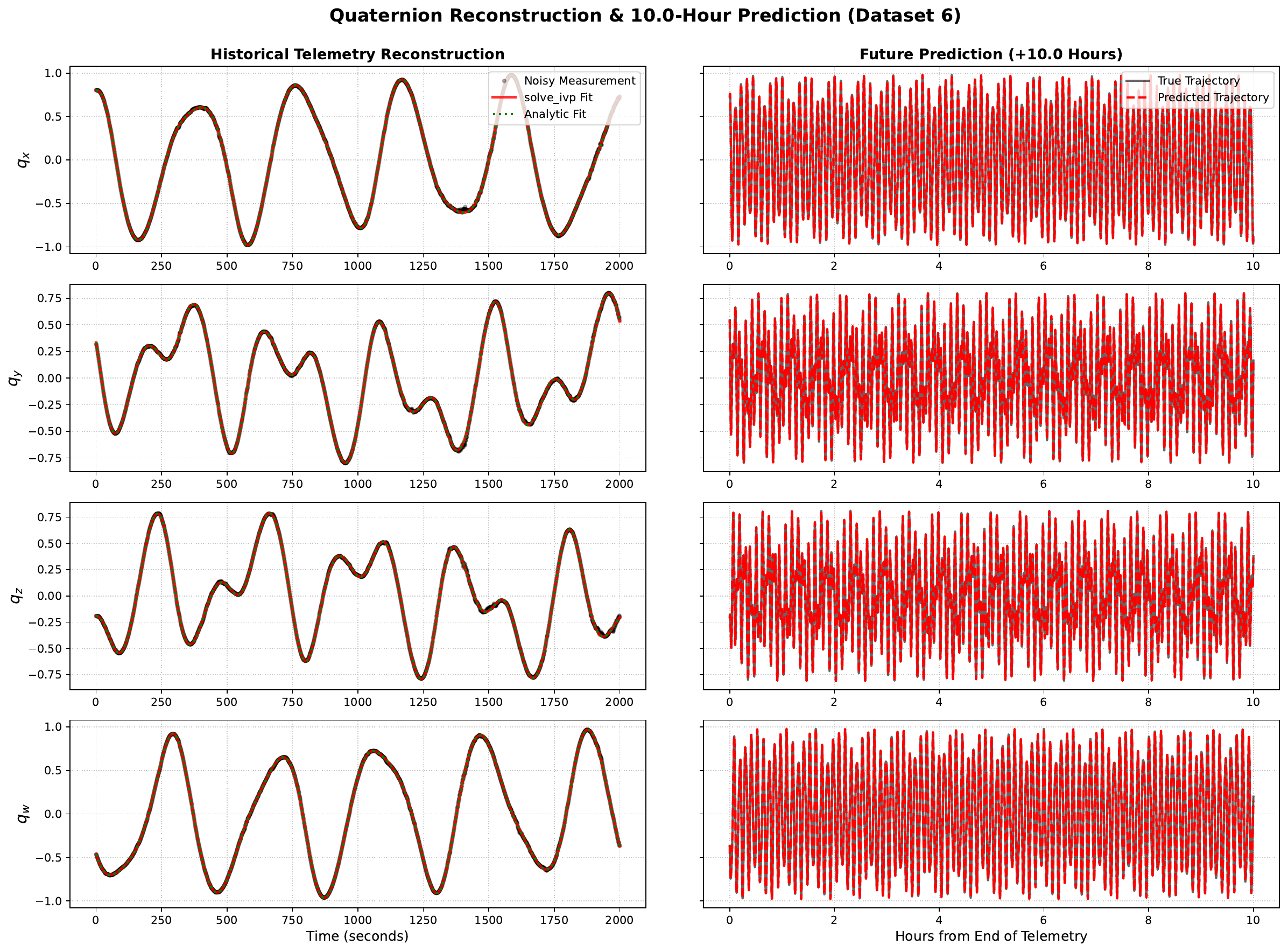}
    \caption{Historical quaternion reconstruction and 10-hour attitude prediction. Left: predicted and observed quaternion histories. Right: long-horizon prediction compared with ground truth, extending 18 times beyond the observation window.}
    \label{fig:param6_historical_future}
\end{figure}

Figure~\ref{fig:param6_historical_future} shows one representative trajectory. The historical quaternion data are reconstructed using the exact Jacobi elliptic functions and a discrete exponential Magnus integrator. The optimization follows Section~\ref{sec:exact_elliptic}, except that the unknown midpoint state $(\boldsymbol{\Omega}_0,\mathbf{q}_0)$ is estimated and then propagated both backward and forward. This midpoint anchoring improves numerical stability when isolated pose-estimation errors occur due to reduced feature visibility. After the inertial parameters are estimated, the trajectory is propagated 10 hours beyond the observation window using a high-order IVP solver and compared with the ground truth. The result demonstrates that the image-derived inertia estimate can support long-horizon attitude prediction for non-cooperative tumbling targets.

\rem{ 
Figure~\ref{fig:param6_prediction_error} quantifies this performance by tracking the absolute attitude divergence over the extended window. Because the initial state and inertia tensor were resolved with high precision during the optimization phase, the predicted trajectory exhibits negligible angular drift, remaining well within the strict operational tolerances required for autonomous deep-space docking maneuvers.

\begin{figure}[htbp]
    \centering
    \includegraphics[width=0.8\textwidth]{step4_mae_error_Param6.pdf}
    \caption{Attitude error evolution over the 10-hour prediction horizon for Dataset Param6. The true angular divergence (in degrees) remains highly constrained, validating the long-term predictive stability of the exact analytic solver for space debris rendezvous.}
    \label{fig:param6_prediction_error}
\end{figure}
} 

\FloatBarrier
\section{Multi-Trajectory and Multi-Arc Inertia Tensor Estimation}
\subsection{Controlled-Noise Multi-Trajectory Evaluation}
\label{sec:controlled_noise_multi}
\begin{table}[htbp]
\caption{Performance comparison of KKT, KKT+EKF, and KKT+Elliptic over $N=100$ realizations. Simulations were performed on an Apple M2 Max laptop.}
\label{tab:mc_comparison_multi_traj}
\centering
\begin{tabular}{lcc}
\hline
\hline 
\textbf{Method} & \textbf{Mean Abs Error} & \textbf{Computational Time (s)} \\
\hline
KKT & 8.22e-05 & 0.0016 \\
KKT + EKF & 1.28e-05 & 6.8997 \\
KKT + Elliptic & 1.12e-06 & 0.2794 \\
\hline
\end{tabular}
\end{table}

\begin{figure}[htbp]
    \centering
    \includegraphics[width=0.6\linewidth]{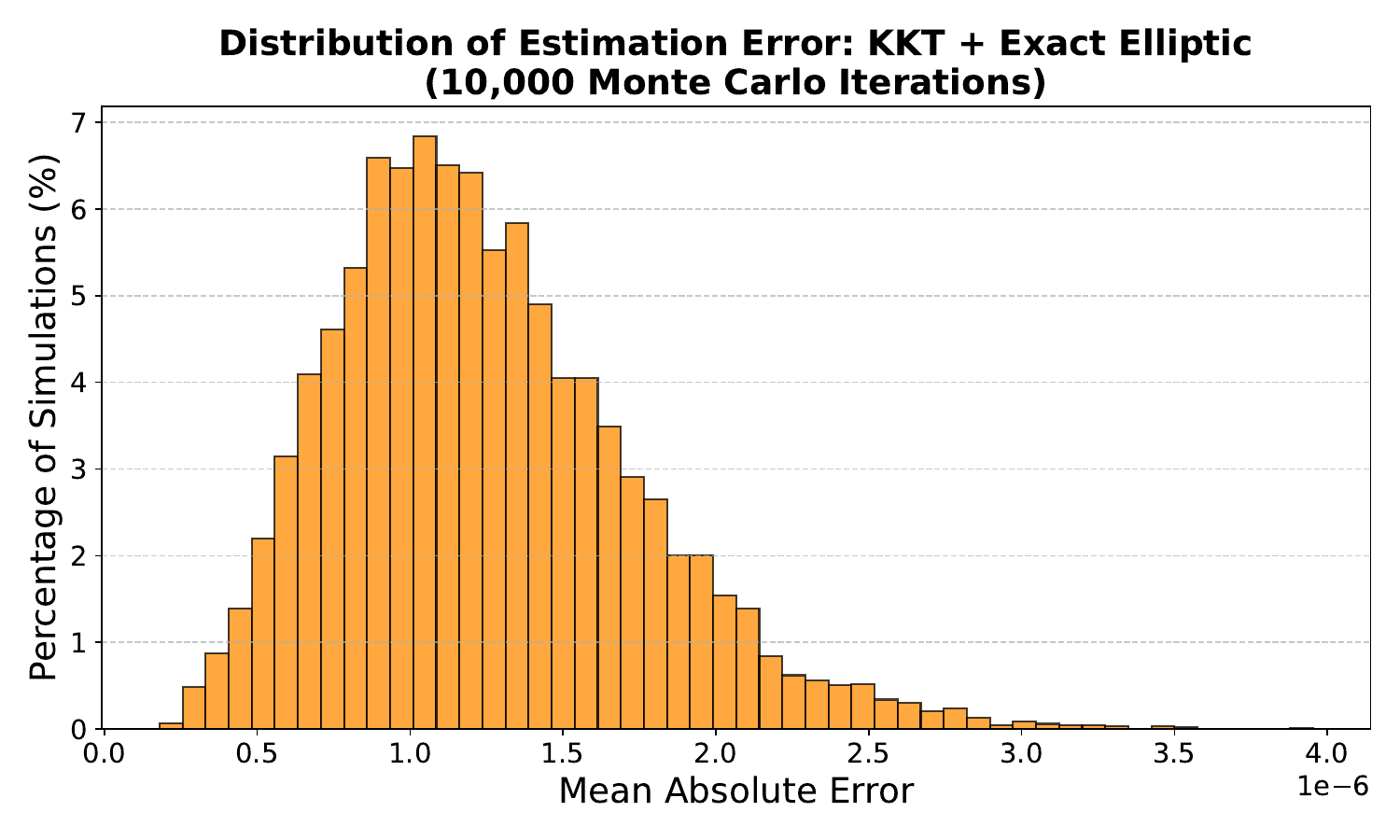}
    \caption{Histogram of multi-trajectory inertia-estimation MAE over 10,000 Monte Carlo realizations. Each case uses three 101-point torque-free trajectories with random initial directions, $\|\boldsymbol{\Omega}_0\|=5~\mathrm{deg/s}$, and 30-arcsecond attitude noise.}
    \label{fig:histogram_multi_traj}
\end{figure}

We next evaluate whether multiple torque-free trajectories of the same object improve inertia-tensor estimation. A single trajectory may not sufficiently excite all rotational modes, especially when the angular velocity is close to a principal inertia axis, leaving some tensor components weakly observable. Combining multiple free-rotation arcs with distinct initial conditions allows the estimator to sample different regions of phase space and reduce these degeneracies.

This section compares three approaches: the KKT initialization alone (KKT), an Extended Kalman Filter refinement initialized from the KKT estimate (KKT+EKF), and the proposed exact elliptic-shooting refinement also initialized from the KKT estimate (KKT+Elliptic). The multi-trajectory setting consists of $K=3$ independent torque-free trajectories of the same satellite, representing passive intervals between maneuvers. Each trajectory is initialized with a randomly chosen angular-velocity direction of magnitude $5~\mathrm{deg/s}$, sampled for 101 points at 1-second intervals, and perturbed by random attitude rotations using the noise model in Section~\ref{sec:data_generation}.

The continuous dynamics are integrated to generate attitude and angular-rate histories, and the experiment is repeated over $N_{\text{sim}}=100$ Monte Carlo trials. For each trial, the three trajectories are jointly used to estimate the normalized inertia tensor with $\operatorname{tr}(\mathbf{I})=1$. Performance is evaluated using the mean absolute error (MAE) of the tensor components. As shown in Table~\ref{tab:mc_comparison_multi_traj}, KKT+Elliptic provides substantially lower error than the baseline methods while remaining, on average, about 20 times faster than KKT+EKF, despite estimating additional initial-state variables $(\boldsymbol{\Omega}_0,\mathbf{q}_0)$ for each trajectory. The corresponding error distribution is shown in Fig.~\ref{fig:histogram_multi_traj}.

\subsection{Image-Based Multi-Arc Evaluation}
\FloatBarrier 

The proposed inertia tensor estimation method is evaluated on the multi-arc RPO dataset using the image-based pose estimates obtained in Section~\ref{subsec:pose_estimation_results}. Each sequence contains two impulsive changes in angular velocity, producing three short torque-free attitude arcs. As described in Section~\ref{subsec:datasets}, the dataset is stratified by three initial angular-velocity regimes (low, medium, and high), and four maneuver modes (SS, SL, LS, and LL). The resulting multi-arc RPO dataset consists of 120 sequences, with each sequence containing 900 images, corresponding to three arcs with 300 images per arc.

To assess the effect of observation length, shorter subsequences are extracted from each arc using $n\in\{30,\allowbreak50,\allowbreak100,\allowbreak150,\allowbreak200,\allowbreak250,\allowbreak300\}$ images per arc. For each case, the three methods defined in Section~\ref{sec:controlled_noise_multi} are applied to the same image-derived attitude histories. 

\begin{figure}[htbp]
    \begin{subfigure}{0.5\linewidth}
        \centering 
        \includegraphics[height=0.21\textheight,keepaspectratio]{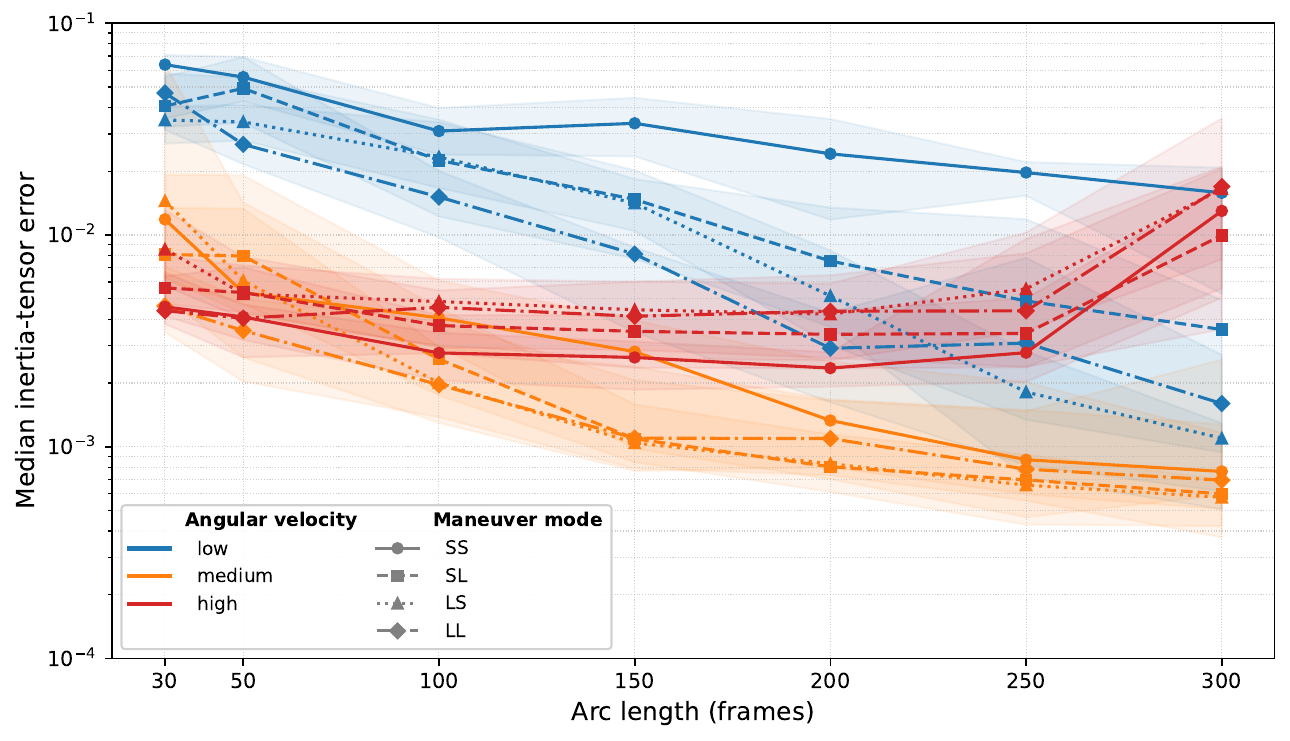}
        \subcaption{}
    \end{subfigure}
    \begin{subfigure}{0.5\linewidth}
        \centering 
        \includegraphics[height=0.21\textheight,keepaspectratio]{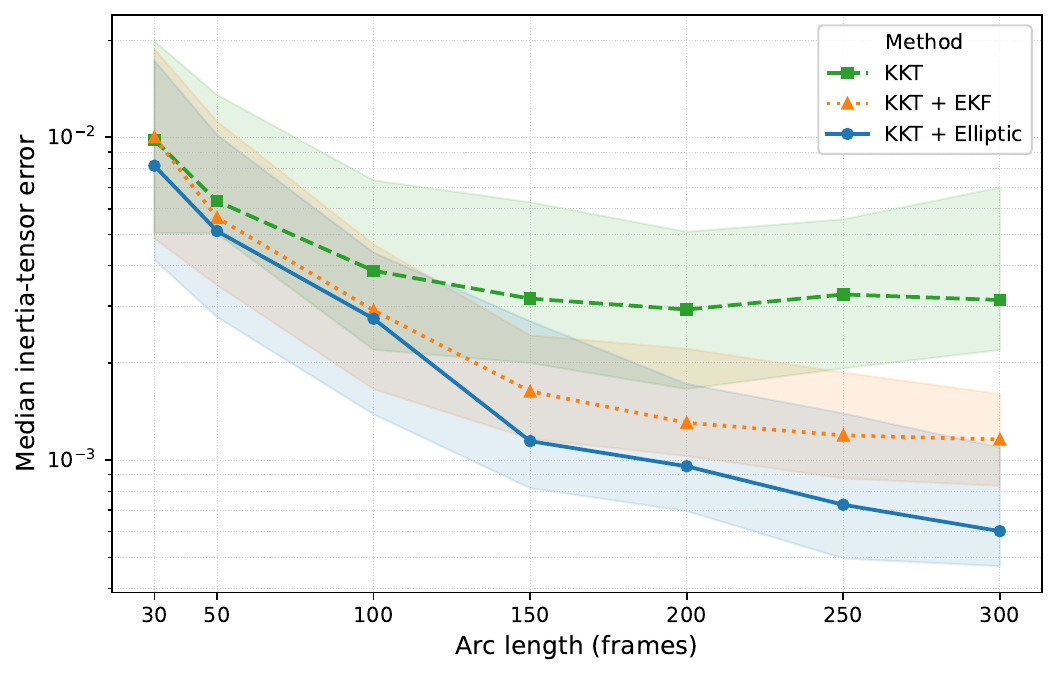}
        \subcaption{}
    \end{subfigure}
    \caption{Convergence of inertia tensor estimation error with observation arc length. (a) KKT+Elliptic results for the 12 angular-velocity-regime/maneuver-mode conditions. (b) Comparison of the KKT, KKT+EKF, and KKT+Elliptic methods for the medium angular-velocity regime, pooled across maneuver modes. Lines denote medians across trial sequences, and shaded bands denote interquartile ranges.}
    \label{fig:maneuver_stratified_function_of_arc_length}
\end{figure}

Figure~\ref{fig:maneuver_stratified_function_of_arc_length} summarizes the inertia-estimation error as a function of observation arc length. The first panel shows KKT+Elliptic results for all 12 stratified conditions. The largest errors occur in the low angular-velocity regime, particularly for short arcs, because the attitude history remains close to constant-rate rotation and provides limited excitation. This limitation is most pronounced for the SS maneuver mode, where both the initial angular velocity and maneuver-induced angular-velocity changes are small.

For the low- and medium-angular-velocity regimes, increasing the arc length generally reduces the estimation error. This trend is less monotonic in the high angular-velocity regime. Although low angular velocity provides insufficient excitation, excessively rapid rotation can also degrade performance because the image cadence is fixed at $1~\mathrm{s}$, reducing the effective temporal resolution of the sampled attitude history. Overall, the medium angular-velocity regime provides the best balance between dynamical excitation and temporal sampling, yielding the lowest errors for $n\geq 100$.

The second panel compares KKT, KKT+EKF, and KKT+Elliptic for the medium angular-velocity regime, pooling results across all maneuver modes. Across all tested arc lengths, KKT+Elliptic achieves lower median inertia-tensor error than both the KKT initialization alone and the KKT+EKF refinement.

\FloatBarrier
\section{Conclusions}

This paper presented an attitude-only framework for estimating the normalized inertia tensor of a spacecraft from torque-free rotational motion. The proposed KKT+Elliptic method uses a KKT-based linear solution as an initial estimate and refines it by minimizing orientation mismatch using the exact elliptic solution of Euler’s equations combined with a Magnus-expansion-based quaternion mapping. The method requires only orientation histories and does not require gyroscope measurements or known control torques.

Using synthetic attitude histories with controlled measurement noise, KKT+Elliptic consistently improved the KKT initialization and outperformed the KKT+EKF reference method. In the single-trajectory case, KKT+Elliptic achieved a mean absolute error of $3.56\times10^{-7}$, compared with $2.65\times10^{-5}$ for KKT and $2.27\times10^{-6}$ for KKT+EKF, while reducing computation time from $11.1043$~s for KKT+EKF to $0.1849$~s. In the multi-trajectory case, the corresponding errors were $1.12\times10^{-6}$, $8.22\times10^{-5}$, and $1.28\times10^{-5}$, respectively.

The observability analysis showed that reliable attitude-only inertia estimation requires sufficient rotational excitation. Motions close to principal-axis rotation provide limited information about the inertia tensor. Near the stable maximum- and minimum-inertia axes, the estimation error follows an algebraic degradation trend, whereas near the unstable intermediate axis the degradation is moderated by the separatrix structure of the torque-free rigid-body phase space.

The framework was then evaluated using image-derived attitude measurements in an RPO scenario. Synthetic JASON-1 images were rendered photorealistically, and the target pose was recovered using keypoint detection followed by EPnP with RANSAC. The resulting pose-estimation pipeline achieved median attitude errors of $0.660^\circ$ for the single-arc RPO dataset and $0.964^\circ$ for the multi-arc RPO dataset.

For the image-derived single-arc case, replacing true attitude histories with estimated attitude histories increased the median inertia-tensor error from $2.0\times10^{-4}$ to $6.5\times10^{-4}$. Despite this degradation, the estimated inertia tensor still supported long-horizon attitude prediction, with a median attitude error of $6.7^\circ$ after 10 hours. In the image-derived multi-arc case, estimation accuracy depended strongly on rotational excitation and temporal sampling. Low angular velocities provided weak observability, whereas excessively high angular velocities reduced the effective temporal resolution under the fixed 1~s image cadence. The medium-spin regime gave the most favorable balance, and KKT+Elliptic consistently achieved lower median inertia-tensor error than both KKT and KKT+EKF.

Overall, the results demonstrate that image-based inertia tensor estimation of a non-cooperative spacecraft is feasible when the observed attitude history contains sufficient rotational excitation and the image-derived pose estimates are reliable. The proposed framework provides a path toward estimating mass-property information from passive optical observations for autonomous inspection, debris characterization, and operations involving tumbling or inactive spacecraft.

\subsection*{Data and Code Availability}
The data and code used to generate the numerical results in this paper are available from the corresponding author upon request.
\subsection*{Declaration of Use of AI-Assisted Tools}
The authors used ChatGPT and Gemini 3.5 Pro for language editing, improving the clarity of the exposition, and streamlining and debugging code. All derivations, numerical experiments, code, results, and conclusions were designed by the authors. All AI-assisted outputs were independently checked, verified, and approved by the authors, who take full responsibility for the content of this manuscript.
\subsection*{Acknowledgements}
We are grateful to Dr. A. Sinclair for bringing this problem to our attention, and to Drs. A. Sinclair and T. Griffith for many helpful discussions. We are also grateful to Profs. A. Bloch, C. Eldred, F. Gay-Balmaz, M. Leok, D. Volchenkov, and D. Zenkov. The work of VP was partially supported by the Shelby Foundation of the University of Alabama. The work of DK was partially supported by faculty startup funds provided by the University of Alabama. 

\FloatBarrier
\bibliography{References_satellites}
\appendix
\section{Exact Solution of Torque-Free Euler Equations via Jacobi Elliptic Functions}
\label{app:euler_derivation}

In this appendix, we present the exact analytical solution for the torque-free rotation of a rigid body using Jacobi elliptic functions. Let $I_1, I_2, I_3$ denote the principal moments of inertia, ordered such that $I_1 > I_2 > I_3$. The components of the angular velocity vector along these principal axes are given by $\Omega_1, \Omega_2, \Omega_3$. In the absence of external torques, the governing equations of motion are expressed as:
\begin{align}
    I_1 \dot{\Omega}_1 &= (I_2 - I_3) \Omega_2 \Omega_3 \label{eq:euler1} \\
    I_2 \dot{\Omega}_2 &= (I_3 - I_1) \Omega_3 \Omega_1 \label{eq:euler2} \\
    I_3 \dot{\Omega}_3 &= (I_1 - I_2) \Omega_1 \Omega_2 \label{eq:euler3}
\end{align}

\subsection{Conservation Laws and Algebraic Isolation}
Because the system is isolated from external torques, both the total rotational kinetic energy $E$ and the magnitude of the angular momentum vector $L$ constitute strict constants of motion:
\begin{equation}
    2E (\boldsymbol{\Omega}_0) = I_1 \Omega_1^2 + I_2 \Omega_2^2 + I_3 \Omega_3^2 = \boldsymbol{\Omega}_0 \cdot \mathbb{I} \boldsymbol{\Omega}_0 \, , \quad
    L^2 (\boldsymbol{\Omega}_0) = I_1^2 \Omega_1^2 + I_2^2 \Omega_2^2 + I_3^2 \Omega_3^2 = \left\| \mathbb{I} \boldsymbol{\Omega}_0 \right\|^2 \, . \label{energy_momentum_cons}
\end{equation}
To obtain an exact decoupled expression, we isolate the squared angular velocities $\Omega_1^2$ and $\Omega_3^2$ as functions of $\Omega_2^2$. Multiplying the energy conservation law in \eqref{energy_momentum_cons} by $I_3$ and subtracting it from the angular momentum ($L$) conservation eliminates $\Omega_3^2$, yielding:
\begin{equation}
    L^2 - 2E I_3 = I_1(I_1 - I_3)\Omega_1^2 + I_2(I_2 - I_3)\Omega_2^2 = \rm{const}
\end{equation}
Solving explicitly for $\Omega_1^2$ gives:
\begin{equation}
    \Omega_1^2 = \frac{L^2 - 2E I_3 - I_2(I_2 - I_3)\Omega_2^2}{I_1(I_1 - I_3)} \label{eq:omega1_isolated}
\end{equation}
Symmetrically, multiplying the energy conservation law by $I_1$ and subtracting the momentum conservation eliminates $\Omega_1^2$, isolating $\Omega_3^2$:
\begin{equation}
    \Omega_3^2 = \frac{2E I_1 - L^2 - I_2(I_1 - I_2)\Omega_2^2}{I_3(I_1 - I_3)} \label{eq:omega3_isolated}
\end{equation}

\subsection{Reduction to the Bounded Elliptic Integral Form}
Squaring the second Euler equation, Eq.~\eqref{eq:euler2}, and substituting the expressions from Eq.~\eqref{eq:omega1_isolated} and Eq.~\eqref{eq:omega3_isolated} creates a single decoupled differential equation for $\Omega_2$:
\begin{equation}
    \dot{\Omega}_2^2 = \frac{(I_1 - I_3)^2}{I_2^2} \left[ \frac{L^2 - 2E I_3 - I_2(I_2 - I_3)\Omega_2^2}{I_1(I_1 - I_3)} \right] \left[ \frac{2E I_1 - L^2 - I_2(I_1 - I_2)\Omega_2^2}{I_3(I_1 - I_3)} \right]
\end{equation}
Remember that $E=E(\boldsymbol{\Omega}_0)$ and $L=L(\boldsymbol{\Omega}_0)$ depend on the initial conditions through \eqref{energy_momentum_cons}.
To evaluate this within a defined kinematic regime, we consider the case where rotation occurs primarily about the major axis ($I_1$), meaning the conservation constants satisfy $L^2 > 2E I_2$. Under this constraint, $\Omega_2$ oscillates between bounded limits where the maximum amplitude $A_2$ occurs at the boundary condition $\Omega_3 = 0$:
\begin{equation}
    A_2^2 = \frac{2E I_1 - L^2}{I_2(I_1 - I_2)}. 
\end{equation}
The matching maximum amplitudes for the remaining axes are defined as:
\begin{equation}
    A_1^2 = \frac{L^2 - 2E I_3}{I_1(I_1 - I_3)}, \quad A_3^2 = \frac{2E I_1 - L^2}{I_3(I_1 - I_3)}
\end{equation}
We now transform the system into a dimensionless form by defining the normalized variable $s = \Omega_2 / A_2$. Substituting $s$ into the system dynamics yields the canonical formulation of a Jacobi elliptic integral:
\begin{equation}
    \dot{s}^2 = \lambda^2 (1 - s^2)(1 - k^2 s^2) \label{eq:canonical_elliptic}
\end{equation}
where the fundamental characteristic frequency $\lambda$ and the elliptic modulus $k$ are explicitly governed by the system invariants and mass distributions:
\begin{equation}
    \lambda(\boldsymbol{\Omega}_0) = \sqrt{\frac{(L^2 - 2E I_3)(I_1 - I_2)}{I_1 I_2 I_3}}  \, , \quad 
    k^2(\boldsymbol{\Omega}_0) = \frac{(I_2 - I_3)(2E I_1 - L^2)}{(I_1 - I_2)(L^2 - 2E I_3)}  \, , \quad E = E(\boldsymbol{\Omega}_0), \, \quad L = L(\boldsymbol{\Omega}_0) \, \mbox{via \eqref{energy_momentum_cons}}. 
\label{lam_k_def}
\end{equation}

\subsection{Analytical Solution Vector}
The non-linear first-order differential equation in Eq.~\eqref{eq:canonical_elliptic} maps to the definition of the Jacobi elliptic sine function, $\text{sn}(\tau, k)$, giving $s(t) = \text{sn}(\lambda t, k)$. Re-scaling back to physical angular velocities yields:
\begin{equation}
    \Omega_2(t) = A_2 \text{sn}(\lambda(\boldsymbol{\Omega}_0) t, k)
\end{equation}
Invoking the fundamental trigonometric identities for elliptic functions ($\text{sn}^2 + \text{cn}^2 = 1$ and $k^2 \text{sn}^2 + \text{dn}^2 = 1$), substitution back into Eq.~\eqref{eq:omega1_isolated} and Eq.~\eqref{eq:omega3_isolated} establishes the complete, exact closed-form solution vector for the rigid body system:
\begin{align}
    \Omega_1(t) &= A_1 \text{dn}(\lambda(\boldsymbol{\Omega}_0) t, k(\boldsymbol{\Omega}_0)) \\
    \Omega_3(t) &= A_3 \text{cn}(\lambda(\boldsymbol{\Omega}_0) t, k(\boldsymbol{\Omega}_0))
\end{align}
For alternate initial conditions where the energy-momentum states dictate $L^2 < 2E I_2$, the body rotates primarily about the minor inertia axis ($I_3$). In that operational regime, the modular roles of the $\text{cn}$ and $\text{dn}$ waveforms are inverted relative to a re-indexed parameter $k$.

\appendix
\section{Extended Kalman Filter Formulation}
\label{app:EKF}

To establish a baseline for sequential state and parameter estimation against the proposed exact elliptic method, we implement a continuous-discrete Extended Kalman Filter (EKF). The EKF jointly estimates the kinematic state and the physical inertia parameters by augmenting the state vector.

\subsection{State Definition and System Dynamics}
To enforce the scale invariance constraint $\operatorname{tr}(\mathbb{I}) = 1$, only five independent components of the inertia tensor are estimated. We define the parameter vector $\mathbf{p} = [I_{xx}, I_{yy}, I_{xy}, I_{xz}, I_{yz}]^T \in \mathbb{R}^5$, which implicitly defines $I_{zz} = 1 - I_{xx} - I_{yy}$. The augmented state vector is given by $\mathbf{x} = \begin{bmatrix} \boldsymbol{\omega}^T & \mathbf{q}^T & \mathbf{p}^T \end{bmatrix}^T \in \mathbb{R}^{12}$.

The continuous-time system dynamics, $\dot{\mathbf{x}} = f(\mathbf{x})$, are governed by the coupled Euler equations and quaternion kinematics, while the inertia parameters are modeled as static constants:
\begin{equation}
    f(\mathbf{x}) = 
    \begin{bmatrix}
        \mathbb{I}(\mathbf{p})^{-1} \left( \mathbb{I}(\mathbf{p})\boldsymbol{\omega} \times \boldsymbol{\omega} \right) \\[6pt]
        \frac{1}{2} \boldsymbol{\Omega}(\boldsymbol{\omega}) \mathbf{q} \\[6pt]
        \mathbf{0}_{5 \times 1}
    \end{bmatrix},
\end{equation}
where $\boldsymbol{\Omega}(\boldsymbol{\omega})$ is the standard skew-symmetric quaternion update matrix.

\subsection{Filter Execution}
Because the unforced rigid body dynamics are highly nonlinear, the continuous-time dynamics are integrated numerically using a high-order Dormand-Prince (DOP853) scheme to generate the \textit{a priori} state estimate $\mathbf{x}_{k|k-1}$. The discrete state transition matrix, $\Phi_k \approx \frac{\partial \mathbf{x}_k}{\partial \mathbf{x}_{k-1}}$, is computed using finite-difference approximations along the integrated trajectory. The \textit{a priori} covariance is propagated via:
\begin{equation}
    P_{k|k-1} = \Phi_k P_{k-1|k-1} \Phi_k^T + Q_k,
\end{equation}
where $Q_k$ is the process noise covariance matrix.

The observation model assumes direct, noisy measurements of the satellite attitude, yielding the linear measurement matrix $H = [\mathbf{0}_{4\times3}, I_{4\times4}, \mathbf{0}_{4\times5}]$. Prior to computing the measurement residual, the double-cover ambiguity of the quaternion space is resolved by ensuring $\mathbf{q}_{k|k-1}^T \mathbf{q}_{\text{obs}, k} \geq 0$; if not, $\mathbf{q}_{\text{obs}, k}$ is negated. 

The state and covariance are then updated using the standard EKF equations:
\begin{align}
    \mathbf{y}_k &= \mathbf{q}_{\text{obs},k} - H \mathbf{x}_{k|k-1}, \\
    S_k &= H P_{k|k-1} H^T + R_k, \\
    K_k &= P_{k|k-1} H^T S_k^{-1}, \\
    \mathbf{x}_{k|k} &= \mathbf{x}_{k|k-1} + K_k \mathbf{y}_k, \\
    P_{k|k} &= (I_{12\times12} - K_k H) P_{k|k-1}.
\end{align}
Following the update step, the quaternion partition of $\mathbf{x}_{k|k}$ is explicitly re-normalized to maintain $\mathcal{S}^3$ manifold constraints. 

To prevent the filter from diverging due to local minima—a common pathology in parameter estimation over $\mathrm{SO}(3)$—the initial parameter state $\mathbf{p}_0$ is seeded using the algebraic solution derived from the KKT method.

\end{document}